\documentclass[journal]{IEEEtran}

\usepackage{amssymb,verbatim,bm,color,times,mathtools,multicol,cuted}
\usepackage{ulem}

\newcommand{\matrixstyle}[1]{\mathrm{#1}}

\newcommand{\expect}[1]{\mathbb{E}_{#1}}
\newcommand{\trace}{\mathrm{tr}}

\newtheorem{remark}{Remark}

\usepackage{graphicx}
\usepackage{stfloats}
\usepackage{tikz}
\usepackage{booktabs}
\usepackage{subcaption}
\usepackage{url}
\usetikzlibrary{fit,positioning,automata,backgrounds}

\title{\LARGE \bf Learning On The Job: Zero-Shot Task Execution under \\ Parametric Uncertainty via Trajectory-Parametrized Dual Control}

\author{
  Victor Vantilborgh\textsuperscript{1}$^\spadesuit$, Hrishikesh Sathyanarayan\textsuperscript{2}, Guillaume Crevecoeur\textsuperscript{1},  Ian Abraham\textsuperscript{2,3}  \& Tom Lefebvre\textsuperscript{1} 
		\thanks{$^\spadesuit$ Corresponding author.}
		\thanks{\textsuperscript{1} Department of Electromechanical, Systems and Metal Engineering, Ghent University, Belgium - Email: \{name.surname\}@ugent.be
		}
        \thanks{\textsuperscript{2} Department of Mechanical Engineering, Yale University, USA - Email: hrishi.sathyanarayan@yale.edu 
		}
        \thanks{\textsuperscript{3} School of Electrical and Computer Engineering, University of Sydney, Australia - Email: ian.abraham@sydney.edu.au
		}
	}

\begin{document}

\maketitle
\thispagestyle{empty}
\pagestyle{empty}

\begin{abstract} 
    Model-based control can achieve reliable task performance, but its effectiveness depends on the accuracy of the underlying model.
    Robots operating under model uncertainty must often adapt to previously unseen payloads, objects, and interaction dynamics to complete a task successfully. Conventional approaches typically rely on a dedicated task- and control-agnostic excitation phase for estimating physics parameters, delaying task execution and collecting potentially irrelevant data. 
    In this paper, we instead consider zero-shot task execution under parametric uncertainty, where online model learning and control proceed concurrently during task execution. 
    Our approach formulates reference generation within a dual control framework to produce task-relevant, informative trajectories that reduce parameter uncertainty in directions critical to task success.
    We predefine a feedback policy with an explicit parameter adaptation law and optimize the reference through the resulting adaptive closed-loop dynamics. 
    We propose two formulations: 1) minimizing expected task cost under parameter uncertainty, and 2) minimizing optimality loss, which quantifies the degradation in task performance caused by planning with an incorrect parameter estimate. 
    By actively optimizing for informative trajectories via a natural Fisher information measure, we tightly approach a lower bound on task-relevant parameter uncertainty while simultaneously achieving reliable task execution. 
    Across diverse tasks, controllers, and adaptation laws, targeted exploration enables the identification required for successful task execution, allowing robots to learn relevant physical parameters while completing the task in a single attempt.
\end{abstract}


\section{Introduction}
    Robots increasingly perform manipulation tasks in manufacturing, logistics, and service applications, where they must interact with diverse objects whose physical properties are often unknown or only partially known.
    Many contact-rich tasks require reliable execution from their very first interactions with unseen objects or payloads. Yet, the physical properties that govern task success, such as mass, inertia, friction, geometry, or contact dynamics, may be unknown a priori and difficult to infer without intentional interaction. Collecting large datasets for every manipulation setting is often impractical~\cite{openx2024, andrychowicz2020learning}, particularly when the relevant dynamics can only be observed in situ. Moreover, identifying every unknown model parameter may be unnecessary, since some uncertainties have little influence on task success. Enabling robots to actively gather informative data to improve task-sensitive model precision while simultaneously adapting their actions towards task completion remains a fundamental challenge.

    Reinforcement learning (RL) can learn complex behaviors without an explicit model, but typically requires many interactions and resets, task-specific reward functions that are difficult to obtain outside curated environments, and may transfer poorly when physical parameters change. 
    When the structure of the dynamics is known and uncertainty is concentrated in a small number of physical parameters, online parameter identification offers a more data-efficient alternative while retaining the structure and safety properties of model-based control. 
    Nevertheless, model-based controllers can degrade substantially under parameter uncertainty, with model mismatch affecting tracking performance and safety-critical functions such as collision detection~\cite{gaz2017payload}.

    Accurate model-based control therefore requires physical models that remain consistent with the system observed through sensory feedback. 
    This has motivated approaches that separate identification from task execution: first excite the system and estimate its parameters~\cite{Hrishi2025, memmel2024asid,YuY-RSS-26,gordon2026active}, then execute the task using the identified model \cite{zhang2025provably,Shi2025,Barcelos-RSS-21}. 
    However, such separation is impractical when environments change frequently or where exhaustive exploration is impractical. 
    Moreover, the excitation trajectory is typically designed independently of the used controller used, and conventional experiment-design objectives are \textit{task-agnostic}, potentially spending effort identifying parameters that have little relevance to the control objective.

\begin{figure*}
        \centering
        \includegraphics[width=\linewidth]{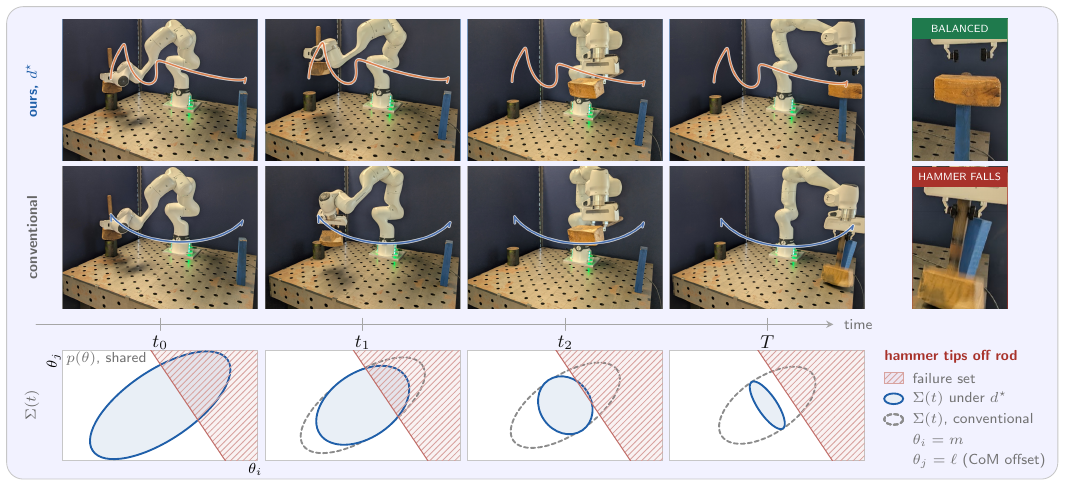}
        \caption{Comparison of dual control and conventional trajectory planning. The Franka FR3 transports an object with unknown mass distribution and balances it on a narrow tower. Top: our dual control optimized trajectory $d^\star$; middle: a conventional task-only reference, with the corresponding end-effector paths overlaid. Bottom: schematic evolution of the parameter covariance $\Sigma(t)$. Under $d^\star$, the task-executing motion also provides task-relevant excitation, reducing uncertainty primarily along parameters that affect task performance, such as the center of mass (CoM), while providing little excitation for parameters such as the moment of inertia that have limited influence on the task. The conventional reference leaves the task-relevant uncertainty largely unresolved.}
        \label{fig:overview}
    \end{figure*}

    Adaptive control provides a mechanism for updating the controller online based on sensory data, but does not by itself determine which data should be collected. Active learning addresses this complementary problem by selecting informative actions based on uncertainty in the underlying parameters. Combining these ideas allows the robot to adapt its model while deliberately shaping its trajectory to acquire the information needed for task execution. This viewpoint leads naturally to dual control~\cite{Feldbaum1960, mesbah2018stochastic}, where exploration and exploitation are coupled: actions influence both task performance and future model accuracy.
    Exact dual control for nonlinear robotic systems leads to a difficult, generally nonconvex trajectory optimization problem, where the coupling between task performance and information acquisition can produce poor local minima and make optimization highly sensitive to initialization. 
    In this work, we fix the feedback policy and its explicit parameter adaptation law in advance, and optimize only the reference trajectory through the resulting adaptive closed loop. 
    The reference guides the resulting robot trajectory to shape both task performance and informativeness of the data received by the adaptation law.
    As illustrated in Fig.~\ref{fig:overview}, this yields trajectories that deliberately excite the uncertain dynamics where this information is relevant to successful task execution by the adaptive closed-loop system, rather than relying on excitation designed independently of the controller. 

    We propose two formulations for this reference design. The first minimizes the expected task cost under parameter uncertainty, propagating the uncertainty through the adaptive closed loop. 
    The second minimizes \textit{optimality loss}, which quantifies the performance degradation incurred when a task is planned using an incorrect parameter estimate. 
    Both formulations reduce to a nominal task cost augmented by a curvature-weighted parameter covariance and naturally optimize a Fisher information measure without introducing a separate arbitrary exploration objective. Since only the reference is optimized, the framework can incorporate any differentiable controller and adaptation law, producing a reference tailored to their closed-loop dynamics.

    In summary, the contributions of this work are as follows:
    \begin{itemize}
        \item A dual control formulation that directly minimizes the expected task cost under parameter uncertainty through the adaptive closed-loop dynamics, 
        \item An alternative optimality-loss formulation that focuses exploration on parameter directions relevant to task performance.
        \item Validation on a variety of robotic manipulation tasks under parametric uncertainty.
    \end{itemize}

    The remainder of the paper is organized as follows: Section \ref{sec:problem} formalizes the problem statement,  Section \ref{sec:related} discusses related work and Section \ref{sec:method} presents the proposed methodology. Section \ref{sec:results} presents the results, followed by the conclusion in Section \ref{sec:conclusion}.  

    \section{Problem formulation}\label{sec:problem}
	Consider the following control system
	\begin{equation}
		\label{eq:sys0}
		\dot{x} = f(x,u,\theta)
	\end{equation}
	where $x\in\mathbb{R}^{n_x}$ denotes the system state, $u\in\mathbb{R}^{n_u}$ the control input and $\theta\in\mathbb{R}^{n_\theta}$ an unknown set of parameters. The parameter $\theta$ belongs to the admissible parameter set $\Theta$ with distribution $\theta \sim  \mathcal{P}$. The function, $f\in\mathbb{R}^{n_x}$, represents the system dynamics.
	
	Our goal is to solve the following optimal control problem
	\begin{equation}
		\label{eq:dualobj}
		\min_u \expect{\theta \sim \mathcal{P}}[J] \text{ s.t. } (\ref{eq:sys0})
	\end{equation}
	where the cost function $J$ is defined as
	\begin{equation}\label{eq:cost}
		J = m(x(T)) + \int_0^T l(t,x(t),u(t))\text{d}t
	\end{equation}
	and $l\in\mathbb{R}$ and $m\in\mathbb{R}$ denote a running and terminal cost.

	This formulation requires the controller to achieve two objectives simultaneously: (i) execute the task optimally, and (ii) identify parameters, but only insofar as this improves task performance. We emphasize this construction grants the control system only a single try to execute the task optimally, where learning and control must occur concurrently. This problem formulation is in stark contrast with state-of-the-art approaches that allow a dedicated identification phase \cite{zhang2025provably}.
    
    As such, problem (\ref{eq:dualobj}) can be identified as a dual control problem \cite{Feldbaum1960}.

    \section{Related Work}\label{sec:related}
    A wide range of strategies have been developed to address the problem posed in Section \ref{sec:problem}, including optimal experiment design, robust and adaptive control, and dual control. 

    \textbf{Optimal Experiment Design.} Optimal experiment design (OED) originates from Fisher's seminal work in statistics~\cite{Fisher1925} and has been widely adopted in system identification to design input trajectories that maximize information for parameter estimation~\cite{Federov2010, Kiefer1959}. The Fisher information matrix $\mathcal{I}(\theta;u)$ quantifies the information acquired under an input trajectory, with scalar criteria such as A-, D-, and E-optimality commonly used to define the design objective~\cite{huan2024optimal}. However, the optimal design depends on the unknown parameters themselves, motivating robust OED approaches that account for parameter uncertainty~\cite{Hagg2013} and adaptive OED approaches that update the design as parameter estimates improve~\cite{Hjalmarsson2005,huan2024optimal}. These approaches primarily optimize information independently of the downstream task, making them task-agnostic. Since a task rarely depends on all parameters equally, other work has shifted toward task-oriented designs that focus only on parameters relevant to control objectives \cite{Jamieson2021, houska2015economic}.  
    
    When applied to control, these criteria typically enter as an additive exploration term,
    \begin{equation}
      \min_u \; \mathbb{E}_{\theta\sim\delta(\theta-\bar\theta)}[J]
            + \mathbb{E}_{\theta\sim\delta(\theta-\bar\theta)}[J_\mathrm{OED}]
      \label{eq:decoupled}
    \end{equation}
    where $\bar\theta$ is a prior parameter estimate. The term $J_\mathrm{OED}$ is designed to incentivize the deterministic control signal, $u$, to generate trajectories that are not only task-relevant but also sufficiently informative for parameter identification.

    \textbf{Learn-then-control.} In robotics, \eqref{eq:decoupled} is often implemented as separate identification and task-execution stages: excite, identify, and only then execute. Recent sim-to-real approaches follow this paradigm by learning exploration policies that maximize Fisher information, collecting real-world data, and subsequently updating the simulator and training the task policy~\cite{memmel2024asid, Shi2025}. Other approaches similarly employ dedicated excitation stages for high-accuracy manipulation~\cite{zhang2025provably} or maximize information about object properties through contact interactions~\cite{Hrishi2025}.
    Such results demonstrate that accurate identification improves downstream control. The separation is nonetheless costly. The excitation phase precedes the task and adds to its duration. More fundamentally, the excitation trajectory is designed independently of the control system that will execute it. Consequently, there is no guarantee that the controller can accurately track the reference under the true $\theta$, or that the trajectory provides informative data under the particular adaptation law used during execution.
    
    \textbf{Adaptive and Robust Control}. Classical strategies for handling uncertainty in robotic manipulation include robust control \cite{spong2020robot}, which guarantees stability under worst-case perturbations at the expense of performance near nominal conditions, and adaptive control, which adapts its control law passively, based on whatever closed-loop response the controller turns out to excite \cite{slotine1991applied, Lee2018}. Distributionally robust formulations replace the single worst case with an ambiguity set of parameter distributions, and have been made tractable for contact-rich manipulation through Stein variational inference~\cite{Hrishi2026}. Model Predictive Control (MPC) offers predictive performance but degrades under poor parameter estimates, while adaptive policy learning fine-tunes a policy online from real-world feedback. These methods act on the controller and leave the reference untouched. A complementary line embeds robustness in the trajectory itself, minimizing the closed-loop sensitivities $\partial x/\partial\theta$ and $\partial u/\partial\theta$ along the reference~\cite{giordano2018, brault2021, wasiela2024, afifi2024}, with validation on quadrotors and a payload-carrying manipulator~\cite{srour2024, srour2025}. It is the mirror image of OED: OED maximizes the sensitivity of the data to $\theta$ with no controller in the loop, whereas sensitivity-aware design minimizes the sensitivity of the closed-loop trajectory to $\theta$ with no estimator in it.

    \textbf{Dual Control and Exploration.} As an extension of adaptive control, a dual controller recognizes that its actions may facilitate acquisition of an improved system model and actively seeks to do so, framing uncertainty-aware control as a balance between exploitation and exploration~\cite{Feldbaum1960}. Both objectives are naturally in conflict and striking the right balance is challenging~\cite{mesbah2018stochastic}. The exact solution is intractable for nonlinear systems, which has motivated approximate dual MPC formulations integrating Fisher-information or Bayesian objectives~\cite{jackson2023dual}, information-seeking MPC over learned ensemble models~\cite{schneider2022}, and Stein variational MPC that jointly infers the dynamics posterior and the control distribution online~\cite{Barcelos-RSS-21}. Reinforcement learning explores similar ideas through curiosity-driven exploration and active inference~\cite{Friston2022, schmidhuber2010formal, houthooft2016vime}, which formalize exploration as greedily maximizing information gain about the environment.
    
    Across these domains, prior work either focuses on information gain independently of the task or designs trajectories without accounting for the controller and adaptation law used during execution. We address these limitations by optimizing the reference trajectory through the complete adaptive closed loop. This couples the effect of parameter uncertainty on task performance to the evolution of the parameter estimate, such that exploration is valuable only insofar as it improves the task. As a result, Fisher information emerges naturally from the closed-loop sensitivity dynamics rather than being introduced as an additional exploration objective, and is weighted by the task cost to emphasize only parameter directions that affect the control objective.

    \section{Dual Control Trajectory Optimization}
	\label{sec:method}

    \begin{figure*}[h]
        \centering
        \includegraphics[width=\linewidth]{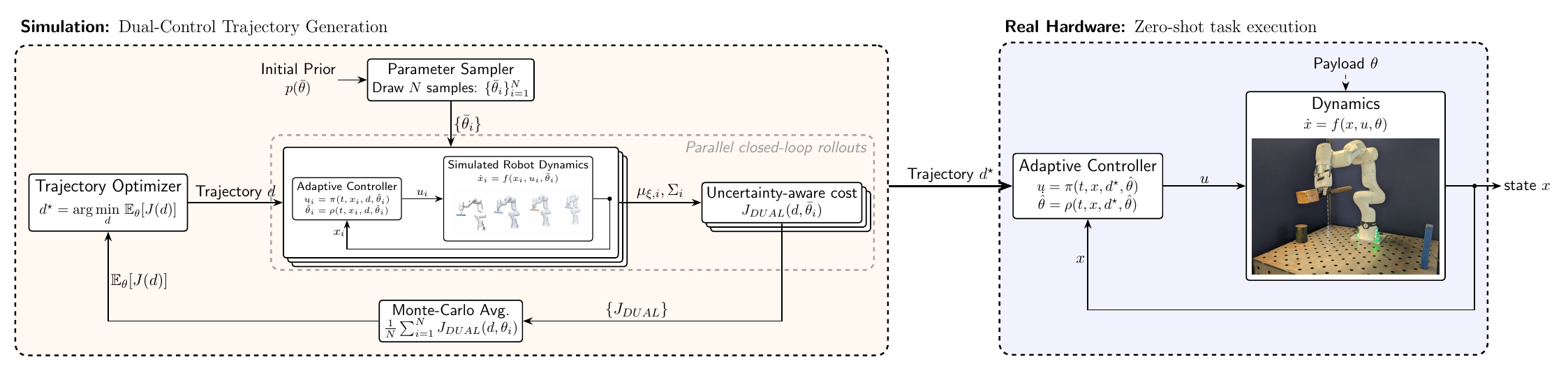}
        \caption{Schematic overview. The reference is optimized offline over samples from the prior, propagated through the closed loop of plant, controller and adaptation law. It is then executed for zero-shot task performance on the real system under the same adaptation control law.}
        \label{fig:schematic_overview}
    \end{figure*}
    
    To account for the flaws identified above we propose to structure the feedback policy, $\pi$, and adaptation scheme, $\rho$, prior to solving problem (\ref{eq:dualobj}). This lowers the complexity of the problem, but not to the extent that is done when addressing the decoupled problem in (\ref{eq:decoupled}). The optimal policy for (\ref{eq:dualobj}) conditions on the observations accumulated during execution and is therefore adaptive by construction. Rather than leaving that mechanism implicit, we impose it, and consider the following modified control system
	\begin{subequations}\label{eq:rcs}
    \vspace{-\abovedisplayskip}
    \begin{align}	
			\dot{x}&= f(x,u,\theta) \\
			u &= \pi(t,x, d, \hat{\theta}) \\
			\dot{\hat{\theta}} &= \rho(t, x,d,\hat{\theta})
		\end{align}	
	\end{subequations}
	Here $d\in\mathbb{R}^{n_d}$ represents a deterministic control design signal, usually the reference signal. The functions $\pi\in\mathbb{R}^{n_u}$, and $\rho\in\mathbb{R}^{n_{\theta}}$ represent the control policy, and adaptation law, respectively. We refer to this reformulation as \textit{Trajectory parametrized dual control}.   
    
	The design, $d$, influences the closed-loop dynamics through two channels. Directly, it shapes the state trajectory through the control policy, $\pi$. Indirectly, it affects the informativeness of the data available for parameter adaptation via $\rho$, and hence the quality of the estimate, $\hat{\theta}$, regardless of the parameter's true value, $\theta$. It is this second channel that a decoupled formulation cannot represent.

    Substituting \eqref{eq:rcs} into \eqref{eq:dualobj} eliminates $u$ and leaves $d$ as the sole degree of freedom. The cost \eqref{eq:cost}, previously a function of the input, becomes a function of the design and the true parameter alone,
    \begin{equation}
    	J(d,\theta) = m\big(x(T)\big)
    	+ \int_{0}^{T}\! l\big(t,x(t),u(t)\big)\,\mathrm{d}t ,
    	\label{eq:clcost}
    \end{equation}
    with $x$ and $u$ generated by \eqref{eq:rcs} for the given $d$ and $\theta$, so that \eqref{eq:dualobj} reduces to the stochastic problem
    \begin{equation}
    	\min_{d}\ \mathbb{E}_{\theta\sim\mathcal{P}}\big[J(d,\theta)\big].
    	\label{eq:stoch}
    \end{equation}
    The decision variable is no longer an open-loop input but a design that will be executed by a known, adapting controller. This is what separates \eqref{eq:stoch} from \eqref{eq:dualobj}.
     
    In this work we consider two active learning objectives to determine the control design $d$. They differ in how the effect of the parameter uncertainty on \eqref{eq:stoch} is evaluated. Fig. \ref{fig:schematic_overview} shows a schematic overview.
    
	\subsection{Robust optimization}\label{sec:RO}
    The straightforward approach is to approximate the expectation in
    \eqref{eq:stoch} directly. The optimization is thereby made aware of the parameter estimation scheme inherent to the control, and the design $d$ is driven to excite the system in such a way that the influence of the initial parameter uncertainty on the outcome of the experiment is reduced.
   
    The expectation is taken about the mean $\bar{\theta}$ of the true distribution of $\theta$, which allows simulation of the closed-loop dynamics during trajectory optimization. Let $\xi=(x,\hat\theta)\in\mathbb{R}^{n_\xi}$ denote the augmented state of
    \eqref{eq:rcs}, written compactly as
    \begin{equation}
    	\dot\xi = g(t,\xi,\theta,d),
    	\qquad \xi(0) = (x_0,\hat\theta_0),
    	\label{eq:aug}
    \end{equation}
    and absorb the policy into the running cost,
    \begin{equation}
    	\ell(t,\xi,d) \coloneqq l\big(t,x,\pi(t,x,d,\hat\theta)\big),
    	\label{eq:runcost}
    \end{equation}
    so that the control no longer appears explicitly and the closed-loop cost
    \eqref{eq:clcost} is a functional of $\xi$ alone. The nominal trajectory and the covariance of $\xi$ then obey
    \begin{subequations}
    	\label{eq:prop}
    	\begin{align}
    		\dot\mu_\xi &= g(t,\mu_\xi,\bar\theta,d), \\
    		\dot\Sigma  &= \mathrm{G}\Sigma + \Sigma\mathrm{G}^{\top},
    	\end{align}
    \end{subequations}
    with
    \begin{equation}
    	\Sigma = \begin{pmatrix}\Sigma_{\xi\xi} & \Sigma_{\xi\theta}\\[1pt]
    	                        \Sigma_{\theta\xi} & Q\end{pmatrix},
    	\quad
    	\mathrm{G} = \begin{pmatrix}\matrixstyle{A} & \matrixstyle{B}\\ 0 & 0\end{pmatrix},
    	\quad
    	\begin{aligned}
    		\matrixstyle{A} &= \tfrac{\partial g}{\partial\xi}\big|_{\mu_\xi,\bar\theta},\\
    		\matrixstyle{B} &= \tfrac{\partial g}{\partial\theta}\big|_{\mu_\xi,\bar\theta},
    	\end{aligned}
    	\label{eq:propblocks}
    \end{equation}
    where $Q$ is the prior covariance of $\theta$; since $\xi(0)$ is deterministic, all initial uncertainty sits in the parameter block. The lower block row of $G$ vanishes because $\theta$ is unknown but fixed, so $\dot\Sigma_{\theta\theta}=0$ and that block stays at $Q$.  What contracts as the estimator learns is the error $\hat\theta-\theta$, recovered from the remaining blocks. More advanced parametric uncertainty propagation schemes, such as polynomial chaos expansions, may be substituted here, though these tend to suffer from the curse of dimensionality.
    
    We can then consider a second-order approximation of the expectation in (\ref{eq:stoch})
    \begin{equation}
		\label{eq:stoch_approx}
		\begin{multlined}
			\expect{\theta\sim p(\bar{\theta})}\left[J(d)\right] = \\
			\begin{aligned}
				&= \expect{\theta\sim p(\bar{\theta})}\left[m(x(T)) + \int_{0}^T \ell(t,\xi(t),d)\text{d}t\right] \\
				&\begin{multlined}
					\approx m(\mu_x(T)) +
					\int_{0}^T \ell(t,\mu_\xi(t),d)\text{d}t +  \\
					 \tfrac{1}{2}\trace\left(\Sigma_{xx}(T)M\right) +  \tfrac{1}{2}\trace \left( \int_{0}^T \Sigma_{\xi\xi}(t) \matrixstyle{L}(t)\text{d}t \right) 
				\end{multlined}\\
				&= \expect{\theta\sim \delta(\theta-\bar{\theta})}[J(d)] + \expect{\theta\sim \delta(\theta-\bar{\theta})}[J_{\text{OED},1}(d)] \\
				&= \expect{\theta\sim \delta(\theta-\bar{\theta})}[J_{\text{DUAL},1}(d)]
			\end{aligned}
		\end{multlined}
	\end{equation}
	where $\matrixstyle{M} = \nabla_x^2 m|_{\mu_x(T)}$ and $\matrixstyle{L} = \nabla_{\xi}^2 \ell|_{\mu_\xi(t)}$. The term $J_{\mathrm{OED},1}$ accounts for the parameter uncertainty and reasons about the Fisher information of the experiment. Unlike classical OED it does so through the estimation law that is actually used, rather than through an assumed optimal estimator. Furthermore the information is scaled by the problem curvature $\mathrm{L}$, so that only those directions of $\Sigma_{\xi\xi}$ which affect the task are actively minimized.

    \begin{remark}
        Note that setting $\rho\equiv 0$ removes $\hat\theta$ from $\xi$ and leaves the open-loop parametric sensitivity of the state. Objective \eqref{eq:stoch_approx} then reduces to sensitivity-aware reference design in the sense of \cite{srour2025}.
    \end{remark}
		
    To limit the coarseness of the second-order approximation of the expectation, which is only valid when the variance of $p(\bar{\theta})$ is small, we use a Gaussian Mixture Model (GMM) to represent $p(\bar{\theta})$ when
    \begin{equation}\label{eq:gmm}
        p(\bar{\theta}) = \sum_i c_i \mathcal{N}(\overline{\theta}_i,Q_i)
    \end{equation}
    where $c_i$ are the coefficients of the GMM components with $\sum_i c_i = 1$. 
    As a result equation (\ref{eq:stoch}) is no longer approximated as in (\ref{eq:stoch_approx}) but by the weighted average
    \begin{equation}
        \expect{p(\bar{\theta})}\left[J(d)\right] = \sum_i c_i \expect{\theta\sim\delta(\theta-\bar{\theta}_i)}[J_{\text{DUAL,1}}(d)]
    \end{equation}
    
    with $\mu_\xi$ and $\Sigma$ propagated from $\theta\sim\mathcal{N}(\theta_i,Q_i)$ rather than from the global prior. This prevents the adaptive controller’s initial parameter estimate from coinciding with the simulated true dynamics, thereby enforcing realistic mismatch and ensuring that parameter adaptation contributes meaningfully to the closed-loop behavior during the reference trajectory optimization.

    \subsection{Optimality loss optimization}
    The approximation \eqref{eq:stoch_approx} quantifies the effect of parameter uncertainty by simulation: the uncertainty on $\theta$ is propagated through the closed loop, and its influence on the trajectory is weighted by the curvature of the cost. We now follow a second route that reaches a criterion of the same form as \eqref{eq:stoch_approx} from a different starting point: the \textit{}{optimality loss} incurred by planning with an incorrect parameter estimate. 
    
    In case we know the value of the true parameter $\theta$ in advance, the solution of (\ref{eq:stoch}) is given by\footnote{Note that this is indeed a deterministic computation even if $\hat{\theta}(0)\neq \theta$.}
    \begin{equation}
        d^\star(\theta) = \arg\min_d J(d, \theta)
    \end{equation}

    In practice, the true parameter is typically unknown and instead we compute a trajectory $d^\star$ using an arbitrary estimate $\bar{\theta}$ of the true parameter $\theta$. Executing this trajectory on the true system yields a suboptimal cost, $J(d^\star(\bar{\theta}),\theta)$. This outcome is suboptimal compared to the true optimal cost, $J(d^\star(\theta),\theta)$, which could only be achieved with perfect knowledge of $\theta$. This performance degradation is referred to as the optimality loss $\Delta(\bar{\theta})$ and is defined as
    \begin{equation}\label{eq:optloss}
        \Delta(\bar{\theta}) = J(d^\star(\bar{\theta}),\theta) - J(d^\star({\theta}),\theta) \geq 0
    \end{equation}

    The loss can be approximated by a second-order Taylor expansion about $\bar\theta=\theta$, with $\tilde\theta=\bar\theta-\theta$,
    \begin{equation}
    	\Delta(\bar{\theta}) \approx \Delta(\theta)
    	+ \tilde{\theta}^\top \frac{\partial \Delta}{\partial \bar{\theta}}(\theta)
    	+ \tfrac{1}{2}\tilde{\theta}^\top \frac{\partial^2 \Delta}{\partial \bar{\theta}^2}(\theta)\,\tilde{\theta} .
    \end{equation}
    The zeroth-order term vanishes by definition. So does the first-order term: only the first argument of \eqref{eq:optloss} depends on $\bar\theta$, so the chain rule gives
    \begin{equation}
    	\frac{\partial \Delta}{\partial \bar{\theta}}(\theta)
    	= \left(\frac{\partial d^\star}{\partial \theta}\right)^{\!\top}
    	  \frac{\partial J}{\partial d}\big(d^\star(\theta),\theta\big) = 0 ,
    \end{equation} 
    Differentiating once more, the stationarity of $d^\star(\theta)$ annihilates the term in $\partial^2 d^\star/\partial\theta^2$, leaving
    \begin{equation}
    	\frac{\partial^2 \Delta}{\partial \bar{\theta}^2}(\theta)
    	= \left(\frac{\partial d^\star}{\partial \theta}\right)^{\!\top}
    	  \frac{\partial^2 J}{\partial d^2}
    	  \left(\frac{\partial d^\star}{\partial \theta}\right) ,
    \end{equation} 
    with both derivatives of $J$ evaluated at $(d^\star(\theta),\theta)$. To evaluate the approximate optimality loss we thus require the sensitivity of $d^\star$ to $\theta$, which follows from the implicit function theorem applied to the stationarity condition $\partial J/\partial d\,(d^\star(\theta),\theta)=0$,
    \begin{equation}
    	\frac{\partial d^\star}{\partial \theta}
    	= - \left(\frac{\partial^2 J}{\partial d^2}\right)^{-1}
    	    \frac{\partial^2 J}{\partial d \partial \theta} .
    	\label{eq:ift}
    \end{equation}
    Substituting, we find
    \begin{equation}
    	\Delta(\bar{\theta}) \approx \tfrac{1}{2}\,\tilde{\theta}^\top \matrixstyle{D}(\bar\theta)\,\tilde{\theta},
    \end{equation}
    where
    \begin{equation}
    	\matrixstyle{D}(\theta) = \left.\left(
    	\frac{\partial^2 J}{\partial \theta \partial d}
    	\left(\frac{\partial^2 J}{\partial d^2}\right)^{-1}
    	\frac{\partial^2 J}{\partial d \partial \theta}
    	\right)\right|_{(d,\theta)=(d^\star(\theta),\theta)} \succeq 0 .
    	\label{eq:D}
    \end{equation}
    As $\theta$ is unknown, $\matrixstyle{D}$ is evaluated at the nominal $\bar\theta$, the substitution perturbs $\Delta$ only at higher order.

    Taking the expected value, $\mathbb{E}[\Delta(\bar\theta)]\approx\tfrac12\operatorname{tr}(D\,\Sigma_{\tilde\theta \tilde\theta})$, we propose

    \begin{equation}
    	J_{\mathrm{OED},2}(d) = \tfrac12\operatorname{tr}\!\big(D(\bar\theta)\,\Sigma_{\tilde\theta \tilde\theta}(T)\big),
    	\label{eq:joed2}
    \end{equation}

    where $\Sigma_{\tilde\theta \tilde\theta}(T)$ is the covariance of the estimation error remaining after the experiment. It is recovered from the off-diagonal blocks of $\Sigma$, not from its $\theta$ block, and satisfies $\Sigma_{\tilde\theta\tilde\theta}(0)=Q$. In line with classical OED it may be replaced by the inverse Fisher information $\mathcal{I}^{-1}(\theta;d)$,
    \begin{equation}
      \mathcal{I}(\theta;d) = \mathbb{E}\!\left[
        \frac{\partial \mathcal{L}(\mathcal{D}\mid\theta)}{\partial \theta}
        \left(\frac{\partial \mathcal{L}(\mathcal{D}\mid\theta)}{\partial \theta}\right)^{\!\top}
      \right],
      \label{eq:fisher}
    \end{equation}
     where $\mathcal{L}$ is the log-likelihood of the data $\mathcal{D}$ collected under the design $d$. 

    In this work, we consider parameter-linear residual models
    \begin{equation}
    r = Y(x,u)\theta + \varepsilon, \qquad \varepsilon\sim\mathcal{N}(0,R),
    \label{eq:measmodel}
    \end{equation}
    where the regressor $Y$ is constructed from the measured states and inputs. This residual constitutes the data $\mathcal{D}$ used in \eqref{eq:fisher} and is also the quantity regressed on by the online estimator. We linearize the regressor noise into $R$, so that the Gaussian likelihood in \eqref{eq:fisher} is an approximation rather than an identity. 
    
     The Cramér–Rao inequality, $\operatorname{cov}(\hat\theta) \succeq \mathcal{I}^{-1}(\theta)$, lower-bounds the covariance of any minimum-variance unbiased estimator and thereby establishes the equivalence. Combining task performance and experimental design gives the single-stage objective
    \begin{equation}
    	J_{\text{DUAL},2}(d) = J(d,\bar\theta) + J_{\text{OED},2}(d),
    	\label{eq:jdual2}
    \end{equation}
    which, under the same mixture decomposition as in \eqref{eq:gmm}, is minimized as
    \begin{equation}
    	\min_d \sum_i c_i\, J_{\text{DUAL},2}(d,\bar\theta_i).
    \end{equation}
     
    \textit{Remark 2:} The weighting matrix $D(\bar\theta)$ highlights the parameter directions that have a large effect on the control objective. Directions in which the design has no leverage, $\partial^2 J/\partial d\,\partial\theta\,v=0$, contribute nothing: knowing $v$ better would not have changed the trajectory. Reducing the uncertainty $\Sigma_{\tilde\theta \tilde\theta}$ along the dominant directions of $D$ therefore removes the effect these parameters have on the original task objective at the least
    cost.

    \textit{Remark 3:} The optimality loss was first proposed in~\cite{houska2015economic} and subsequently used as a cost augmentation in~\cite{feng2020scheme,jackson2023dual}. Those works substitute it into the decoupled problem \eqref{eq:decoupled}, whereas here it is evaluated on the closed loop \eqref{eq:rcs}, which is what makes it comparable to \eqref{eq:stoch_approx}. The original intent was to design an experiment exciting only those dynamics needed to learn task-relevant parameters. Since $\Sigma_{\tilde\theta \tilde\theta}(T)$ measures the uncertainty remaining after the experiment, this is arguably not a suitable criterion when the estimate is used during the same horizon in which it is acquired; a running penalty would be more consistent with the present setting. We retain the terminal form as adopted in~\cite{feng2020scheme,jackson2023dual}, and note that \eqref{eq:stoch_approx} imposes the running weighting automatically.

    \subsection{Relation between the two formulations}
    \label{sec:relation}
    
    Although obtained from different starting points, both objectives share the form
    \begin{equation}
    	J(d,\bar\theta) + \tfrac12\trace\!\big(\matrixstyle{W}\Sigma\big),
    	\label{eq:generic}
    \end{equation}
    a nominal task cost augmented by a curvature weighted by a parameter covariance. For the robust optimization, $W=\partial^2J/\partial\theta^2$ and $\Sigma=Q$. The curvature is never formed explicitly but assembled by propagating $Q$ through the closed loop and weighting the result by $\matrixstyle{L}$ and $\matrixstyle{M}$. For the optimality loss, $\matrixstyle{W}=\matrixstyle{D}$ and $\Sigma=\Sigma_{\tilde\theta \tilde\theta}(T)$, and the curvature is formed directly but only partly: comparing $J(d^\star(\bar\theta),\theta)$ against $J(d^\star(\theta),\theta)$ reflects only the degradation that a different trajectory design would have avoided. What is discarded is the sensitivity that persists even when the design is optimal for the true parameter, since some payloads simply make the task harder.
    
    The two formulations also differ in which factor of \eqref{eq:generic} carries the dependence on $d$. In the robust optimization the covariance is the prior and is therefore fixed, so the design acts entirely through the curvature, reassembled from \eqref{eq:prop} at every iterate. In the optimality loss the curvature is frozen at $\bar\theta$ and identifies which parameter directions matter to the task, while the design acts through $\Sigma_{\tilde\theta \tilde\theta}(T)$, which measures how well those directions are resolved.
    The Fisher information consequently enters implicitly in the first case, as the rate at which the estimation error contracts under an informative design, and explicitly in the second through \eqref{eq:fisher}. For a linear--Gaussian estimator the correspondence is exact: in information form, the propagation (\ref{eq:prop}) contracts the error covariance at rate $Y^{\!\top} R^{-1} Y$, with $Y$ and $R$ as in \eqref{eq:measmodel}, this is precisely the Fisher information rate of the experiment. For other adaptation laws the correspondence holds only approximately. The first requires $\rho$ to be simulated inside the optimization and its weighting remains valid as the design moves. The second needs only a model of the estimator accuracy, but its weighting is linearized about the nominal design. In both cases it is the weighting that makes the criterion task- and controller-aware.

    \section{Results}\label{sec:results}
    
    \subsection{Cartpole Swing-Up}
    \label{sec:res-cartpole}

     \begin{figure}[b]
        \centering
        \includegraphics[width=\linewidth]{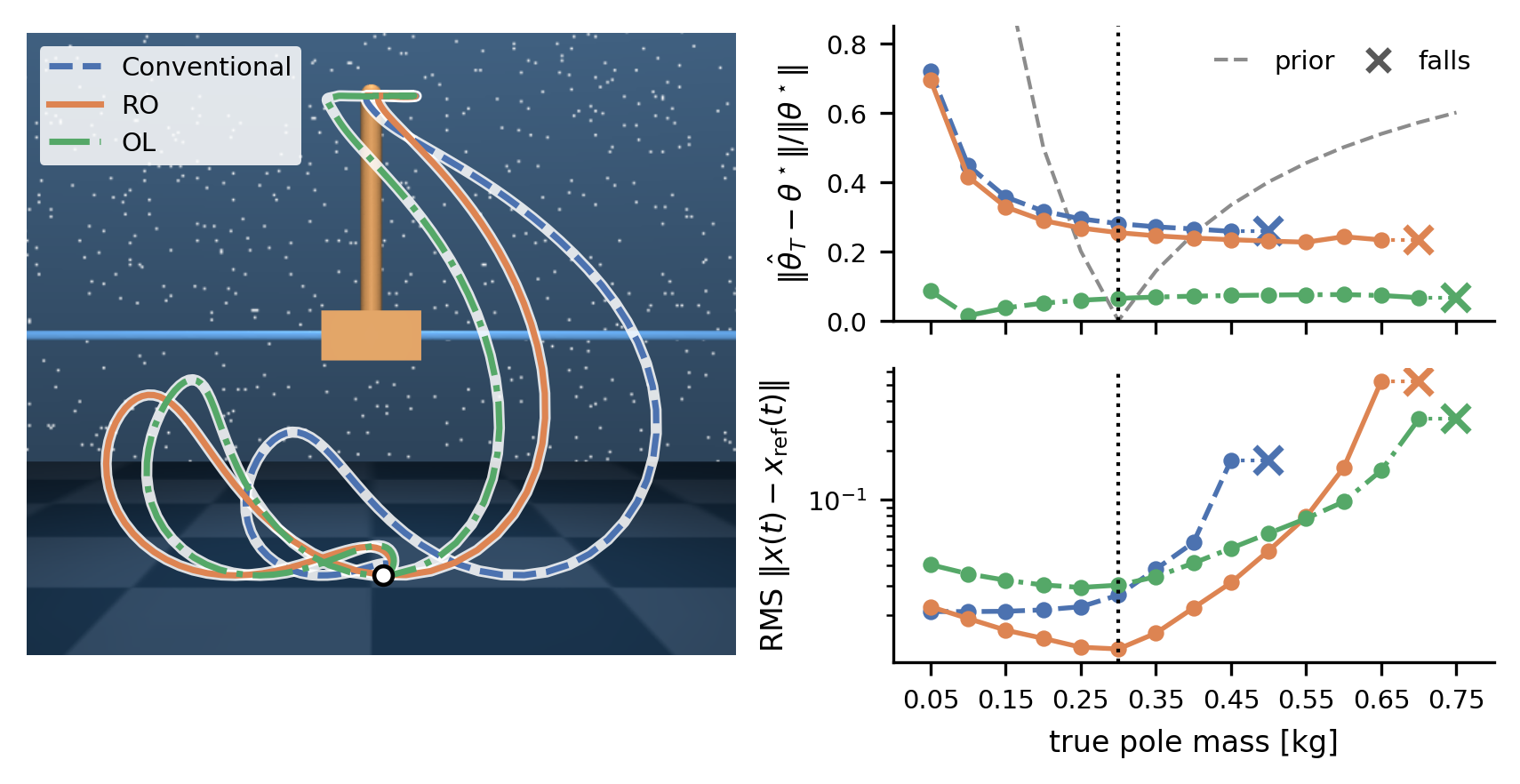}
        \caption{Cartpole results. Left: pole-tip paths of the three references. Right, top: relative terminal identification error $\|\hat\theta_T-\theta^\star\|/\|\theta^\star\|$ against the true pole mass. Right, bottom: closed-loop sensitivity to the parameter mismatch, measured as the RMS deviation of the executed trajectory from its own reference. Each curve spans the masses at which that reference completes the swing-up and ends in a $\times$ at the first mass at which the pole falls.}
        \label{fig:cartpole_combined}
    \end{figure}
    
    We first consider a force-limited cartpole ($|F|\leq 10$~N) that must swing up from the hanging equilibrium and remain upright over a horizon of $T=10$~s. The uncertain parameter is the pole inertia, parametrized as $\theta=[m,h,J]$, where $m$ is the pole mass, $h=m\ell_c$ is its first mass moment about the hinge, and $J$ is its moment of inertia about the hinge. The policy $\pi$ is a time-varying LQR controller~\cite{tedrake_underactuated} that tracks the reference, with its feedforward term recomputed online from $\hat{\theta}$. Parameter adaptation uses recursive least squares (RLS)~\cite{astrom1995adaptive}, and the reference is parametrized by direct collocation~\cite{hargraves1987direct}.
    
    For each method, a single reference is designed using a prior mean $\bar{m}=0.30$~kg and evaluated on $15$ true pole masses uniformly spanning $[0.05,0.75]$~kg. The pole length is sampled independently for each draw, thereby varying $h$ and $J$. The controller, adaptation law, and prior are identical across methods; only the reference differs. This tests whether the designed reference remains effective as the true parameters deviate from the planning prior. The results are summarized in Fig.~\ref{fig:cartpole_combined} and Table~\ref{tab:cartpole}.
    
\begin{figure}[h]
        \centering
        \includegraphics[width=\linewidth]{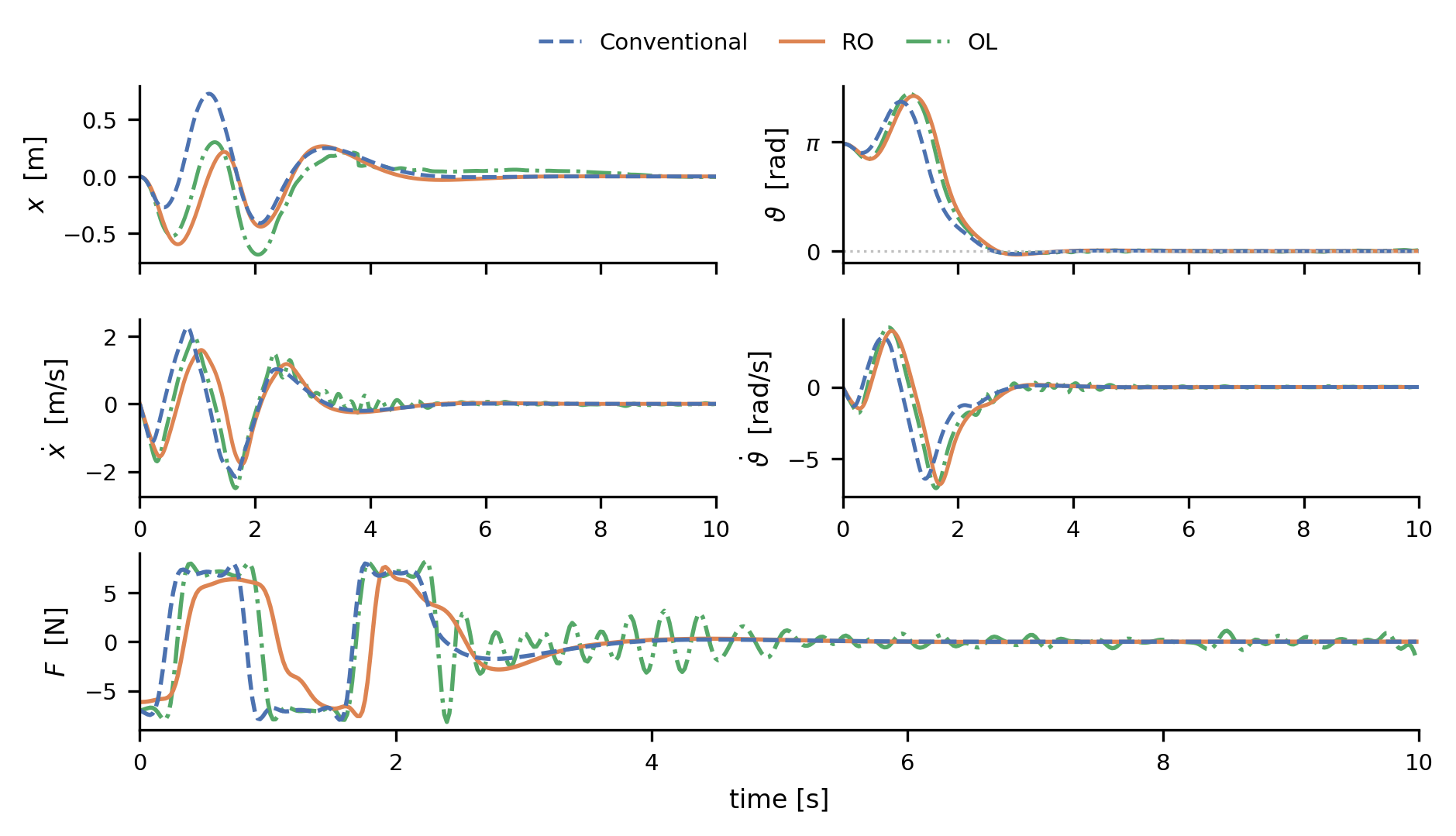}
        \caption{Cartpole references at the planner prior: cart position and velocity, pole angle and rate, and input force. All three perform a two-swing pump; only OL keeps exciting the system once the pole is upright.}
        \label{fig:cartpole_states}
    \end{figure}

    The conventional reference succeeds up to $1.5\times$ the prior mass, failing from $0.50$~kg onwards. RO and OL extend the successful range to $2.2\times$ ($0.65$~kg) and $2.3\times$ ($0.70$~kg), respectively. Fig.~\ref{fig:cartpole_states} shows that at the prior, all three references perform a similar two-swing pump but differ in excitation magnitude and persistence. RO is the least aggressive, with a peak cart velocity of $1.78$~m/s and maximum displacement of $0.59$~m, compared with $2.26$~m/s and $0.72$~m for the conventional reference and $2.50$~m/s and $0.68$~m for OL. OL provides the strongest excitation during swing-up and continues to excite the system after the pole reaches the upright position, whereas the other references settle immediately.
    
    The resulting identification and closed-loop sensitivity are shown in Fig.~\ref{fig:cartpole_combined}. OL maintains a relative terminal parameter error of approximately $7\%$ across the full mass range, compared with approximately $30\%$ for both the conventional and RO references. RO instead substantially reduces sensitivity to parameter mismatch: at $0.45$~kg, the highest mass for which the conventional reference succeeds, its RMS trajectory deviation from the reference is $5.6\times$ smaller. Thus, RO improves robustness primarily through reduced closed-loop sensitivity, whereas OL improves it through more accurate parameter identification.
    
    The two formulations therefore exploit the two mechanisms in \eqref{eq:generic} differently. OL acquires most of its information during swing-up and continues to excite the system during balancing, preventing the estimate from degrading once the pole becomes stationary. RO achieves comparable robustness with substantially less excitation by producing a trajectory that is less sensitive to parameter mismatch.

    \begin{table}[t]
    \centering
    \caption{Cartpole summary over the mass sweep. Success is the fraction of the $15$ true masses at which the swing-up completes; the remaining columns are  means $\pm$ standard deviations over the $9$ masses at which all three references succeed, so the comparison is one-to-one.}
    \label{tab:cartpole}
    \begin{tabular}{lccccc}
    \toprule
     & success & tracking & parameter & input & $t_{\rm up}$ \\
     & [\%] & RMSE [--] & error [--] & cost & [s] \\
    \midrule
    Conv.  & $60$ & $0.045 \pm 0.047$ & $0.356 \pm 0.140$ & $101$ & $2.38$ \\
    RO            & $87$ & $0.018 \pm 0.006$ & $0.329 \pm 0.140$ & $72$  & $2.50$ \\
    OL            & $93$ & $0.036 \pm 0.007$ & $0.058 \pm 0.020$ & $119$ & $2.40$ \\
    \bottomrule
    \end{tabular}
    \end{table}

    \subsection{Quadrotor with a cable-suspended load}
    \label{sec:res-drone}
    Next, we consider a quadrotor carrying a point-mass payload on a slack-free cable. The uncertain parameter is $\theta=[m_L,L]$, comprising payload mass and cable length, drawn from independent Gaussian priors $m_L\sim\mathcal{N}(1.00,\,0.15^2)$~kg and $L\sim\mathcal{N}(0.70,\,0.20^2)$~m. The nominal parameter $\bar{\theta}$ is the prior mean. These parameters are unequally observable: $L$ enters the regressor only through the product $\beta=m_LL$ multiplying the cable angular acceleration, whose contribution vanishes in hover. The estimator therefore works in the linear-in-parameters coordinates $[m_L,\beta]$, with $\hat L=\hat\beta/\hat m_L$. Payload mass can consequently be identified from the thrust required to maintain position, whereas cable length requires excitation of the payload swing. The policy is the geometric controller of~\cite{lee2010geometric}, which tracks the load path by offsetting the vehicle setpoint using the current length estimate, and adaptation uses recursive least squares. The reference is a degree-$5$ B-spline~\cite{biagiotti2008} over a $24$~s horizon.
    
    The task is a $10.5$~m corridor containing a staggered pair of walls, a low tunnel, and a three-sided delivery bay. The final approach requires a fast $0.7$~m lateral move in $1.8$~s, which excites the payload swing and is compensated by a two-impulse input shaper spaced at half the estimated swing period, $\pi\sqrt{\hat L/g}$~\cite{singer1990preshaping}. Since $\omega_n=\sqrt{g/L}$ is independent of $m_L$, the shaper depends directly on the parameter that cannot be identified in hover.
    
\begin{figure}[hb]
        \centering
        \includegraphics[width=\linewidth]{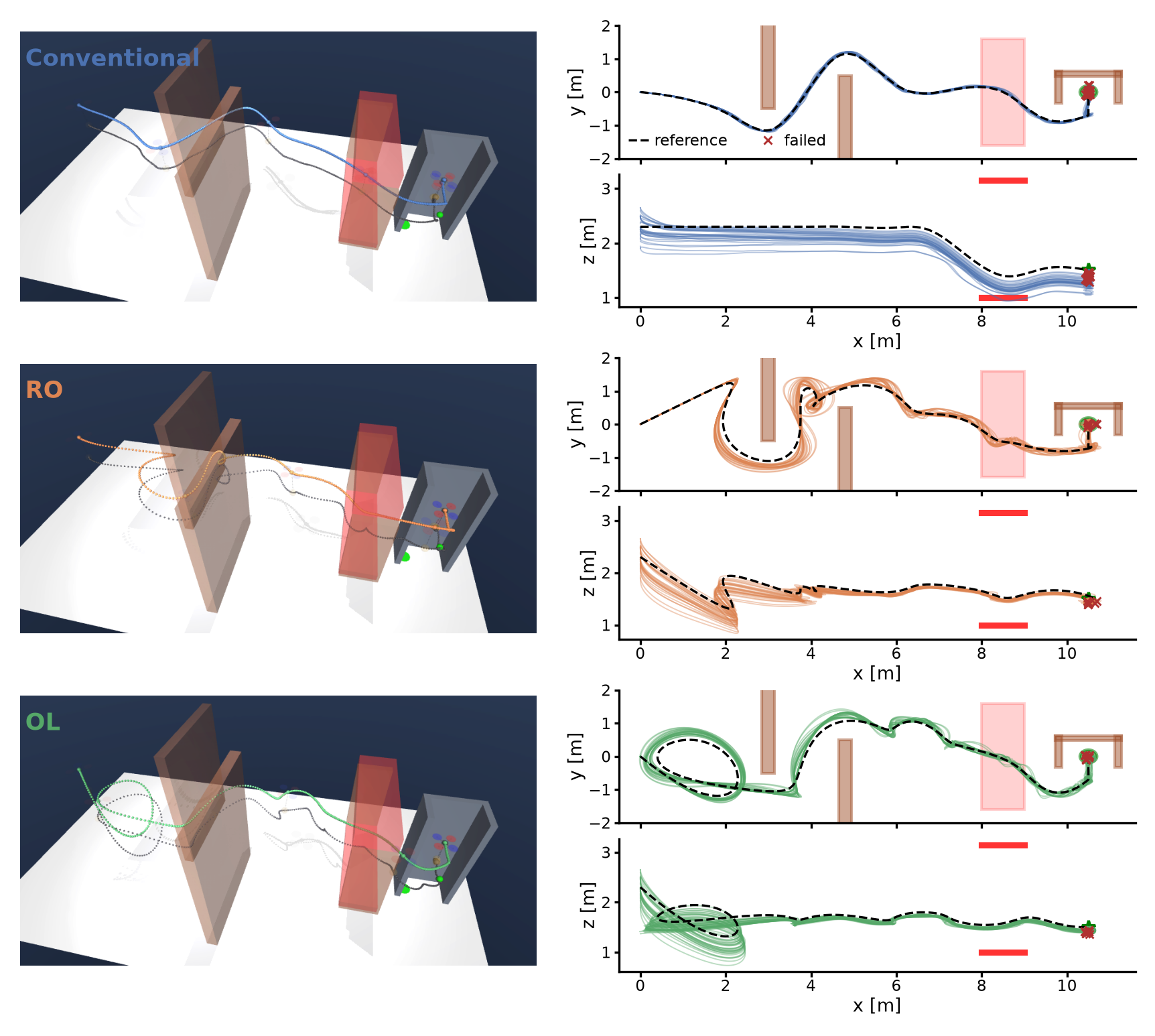}
        \caption{Slung-payload references. Left: renders at the planner prior, vehicle path colored and load path gray. Right: realized load paths over the $30$ payload draws in top ($y$--$x$) and side ($z$--$x$) view; dashed: reference, $\times$: failed delivery.}
        \label{fig:drone_paths}
    \end{figure}

    As shown in Fig.~\ref{fig:drone_paths}, both dual control references concentrate additional excitation in the open section of the course. They increase the peak payload swing to approximately $19^\circ$, compared with $13^\circ$ for the conventional reference, while allowing the swing to decay before the payload enters the walls and tunnel. Both also introduce a small vertical oscillation that further excites the suspended load. Thus, the references acquire the information required for delivery before entering the constrained section of the course.

\begin{figure}[h!]
      \centering
      \begin{subfigure}[t]{0.485\columnwidth}
        \includegraphics[width=\linewidth]{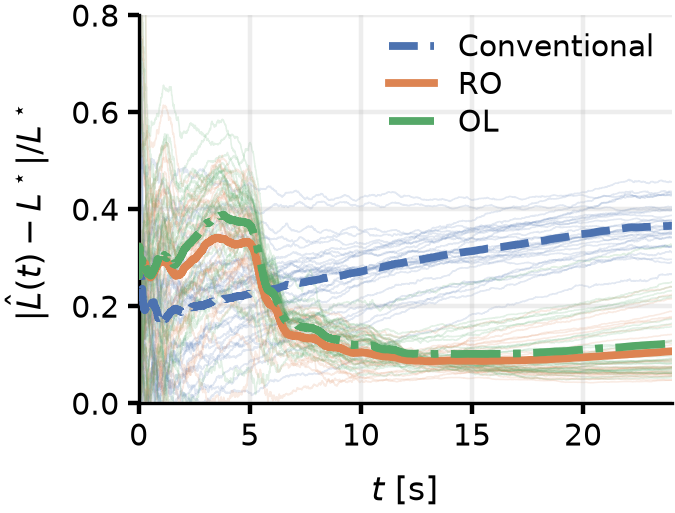}
        \caption{}\label{fig:drone_lerr}
      \end{subfigure}\hfill
      \begin{subfigure}[t]{0.485\columnwidth}
        \includegraphics[width=\linewidth]{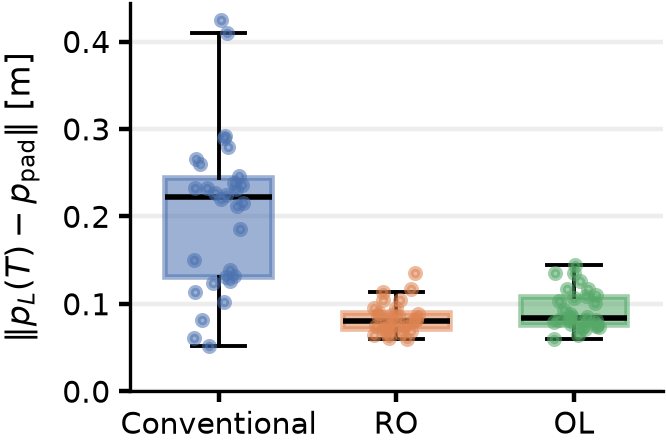}
        \caption{}\label{fig:drone_placement}
      \end{subfigure}
      \vspace{2pt}
      \begin{subfigure}[t]{\columnwidth}
        \includegraphics[width=\linewidth]{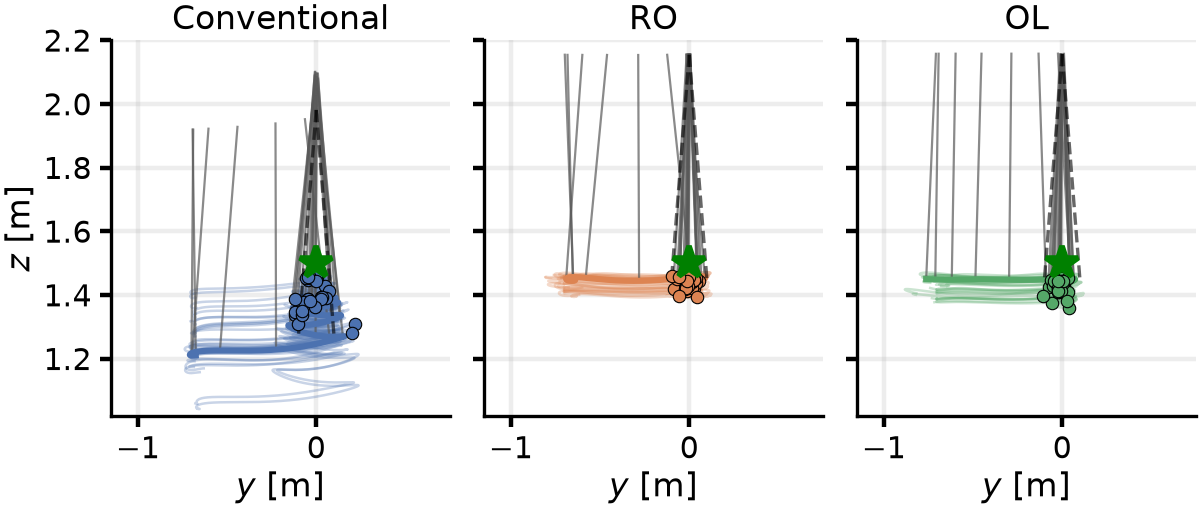}
        \caption{}\label{fig:drone_swing}
      \end{subfigure}
      \caption{Cable-suspended payload delivery over $30$ payload draws.
      (\subref{fig:drone_lerr}) Relative cable-length error; thin lines are individual draws, thick lines the ensemble mean. (\subref{fig:drone_placement}) Terminal placement error of the payload.
      (\subref{fig:drone_swing}) Payload trajectory in the $y$--$z$ plane.}
      \label{fig:drone_results}
    \end{figure}

    This excitation substantially improves cable-length identification (Fig.~\ref{fig:drone_results}a, Table~\ref{tab:drone_results}). The terminal cable-length error is $0.062$~m for RO and $0.072$~m for OL, compared with a prior error of $0.159$~m. The conventional reference provides insufficient excitation and ends with an error of $0.171$~m, slightly worse than the prior. The improved estimate directly benefits delivery: cable-length error affects the vertical payload position through the controller offset $p_d=p_{\mathrm{ref}}+\hat L e_3$ and the timing of the input shaper. The conventional reference consequently produces both tunnel violations and excessive residual swing, resulting in $6/30$ collisions. RO and OL complete all $30$ deliveries without collision, with residual swings of $6.1^\circ$ and $5.6^\circ$, respectively.
    
    RO and OL differ mainly in how much information they acquire and how that information translates into terminal performance. RO obtains the more accurate cable-length estimate, while OL achieves the lower residual swing. Both references concentrate excitation in the unconstrained part of the trajectory, illustrating that information can be acquired without compromising the later delivery phase. The required excitation and clearance constraints are therefore both important to maintaining feasibility while exploiting the available excitation opportunities.
    
    \begin{table}[h]
      \centering
      \caption{Slung-payload delivery, mean,$\pm$,std over $30$ payload draws. Placement error $|p_L(T)-p_\mathrm{pad}|$ at release; swing is the residual amplitude at the drop point. A collision is counted if the payload or vehicle contacts a wall or violates the tunnel floor or ceiling.}
      \label{tab:drone_results}
      \scriptsize
      \setlength{\tabcolsep}{6pt}
      \begin{tabular}{lcccc}
        \toprule
        Method & $|\hat L-L^\star|$ [m] & $\|\Delta p_L\|$ [m]
               & swing [$^\circ$] & collisions \\
        \midrule
        Conventional  & $0.171{\pm}0.086$ & $0.205{\pm}0.090$ & $10.1{\pm}1.9$ & $6/30$ \\
        RO            & $0.062{\pm}0.017$ & $0.084{\pm}0.018$
                      & $6.1{\pm}1.6$ & $0/30$ \\
        OL            & $0.072{\pm}0.024$ & $0.093{\pm}0.023$
                      & $5.6{\pm}1.4$ & $0/30$ \\
        \bottomrule
      \end{tabular}
    \end{table}

    \subsection{Franka FR3: rod balancing task}
    \label{sec:res-rod}

    The final platform is a 7-DoF Franka FR3 that must grasp a rod with unknown mass distribution, transport it across the workspace, and balance it on a narrow tower. Unlike the previous tasks, the terminal target itself depends on $\theta$: successful placement requires the rod's center of mass to lie over the tower. The terminal task cost therefore depends directly on the CoM estimation error, making this parameter direction critical for successful execution.
    
    The uncertain parameter is the rod's full inertial vector
    \begin{equation}
      \theta = \big[\,m,\; c^{\!\top},\; I_{xx},\, I_{yy},\, I_{zz},\,
                    I_{xy},\, I_{yz},\, I_{xz}\,\big]^{\!\top} \in \mathbb{R}^{10},
    \end{equation}
    where $m$ is the rod mass, $c\in\mathbb{R}^3$ is the center of mass in the flange frame, and the remaining six entries are the independent elements of the symmetric inertia tensor about the center of mass. We denote by $c_x$ the component along the rod's long axis, which is the parameter direction most directly relevant to the task. The policy $\pi$ is a Slotine--Li tracking controller~\cite{slotine1991applied} with composite adaptation on the pseudo-inertia manifold, ensuring physically consistent updates~\cite{Lee2018}; the reference is a joint-space B-spline~\cite{biagiotti2008}.

    \begin{figure}[h]
        \centering
        \begin{subfigure}{0.9\linewidth}
            \centering
            \includegraphics[width=\linewidth]{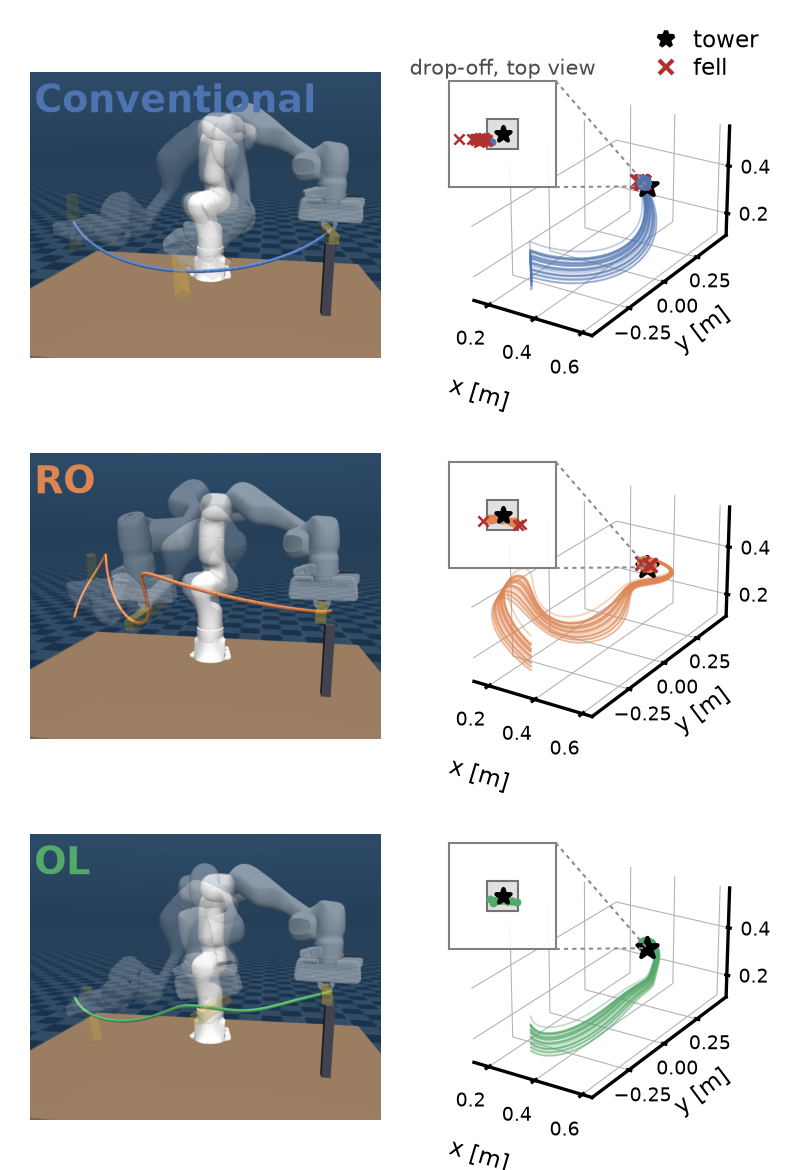}
            \caption{Left: overlay of the reference at the prior, at $t=1.0$, $3.0$, and $4.5$s, with the end-effector path. Right: CoM trajectories for the $30$ sampled rods, with release positions marked ($\times$: subsequent tip) and the tower marked by a star.}
            \label{fig:fr3_3d_plot}
        \end{subfigure}
        \vfill
        \begin{subfigure}{0.8\linewidth}
            \centering
            \includegraphics[width=\linewidth]{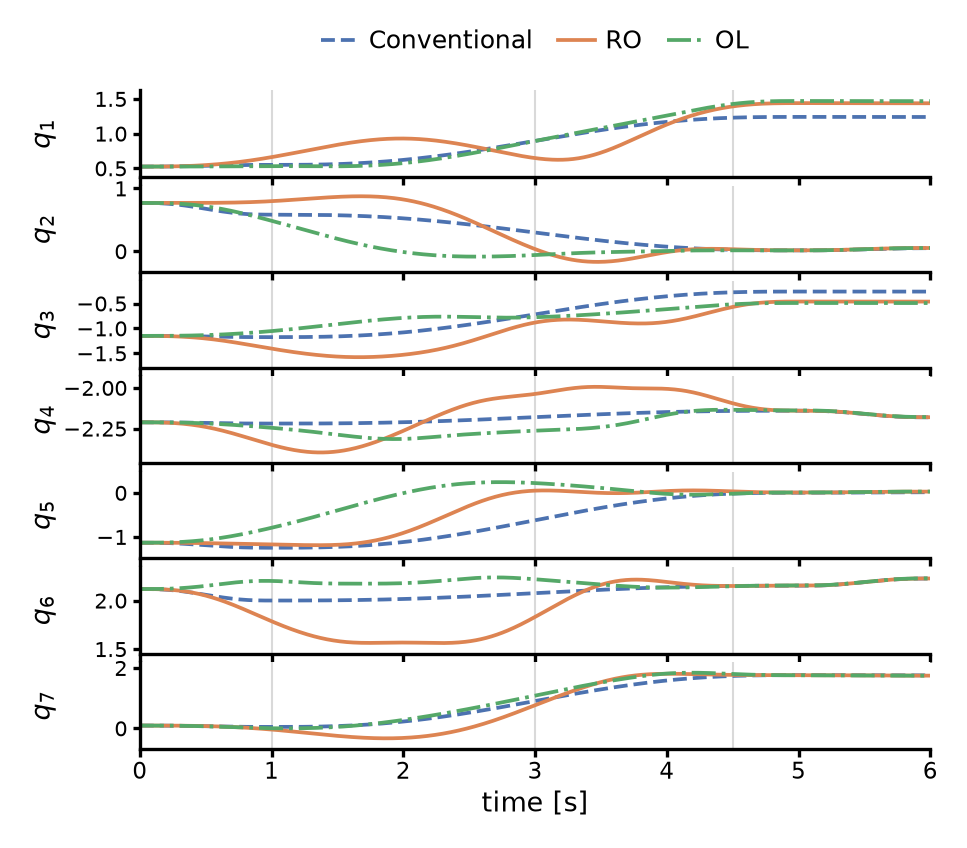}
            \caption{The same three references, joint by joint.}
            \label{fig:joint_plot}
        \end{subfigure}
        \caption{Designed reference trajectories at the planner prior, and the outcomes they produce over $30$ rods drawn from that prior.}
    \label{fig:rod_trajectories}
    \end{figure}
    
    \emph{Simulation.}
    We draw $30$ rods from the prior and execute every reference on every draw. Table~\ref{tab:rod_sim} summarizes the results. The conventional reference balances only $5/30$ rods, with a mean terminal CoM error of $23.5$~mm, exceeding the $15$~mm tower half-width. RO and OL reduce the CoM error to $10.7$ and $6.7$~mm and balance $27/30$ and $30/30$ rods, respectively.
    
    Fig.~\ref{fig:fr3_3d_plot} shows the resulting references and CoM trajectories. The conventional reference moves directly to the tower, whereas both dual control references introduce a lateral excursion that rotates the rod about its grasp point. This additional motion substantially improves the information available to the adaptation law. Because the natural-gradient update depends on the current estimate through the inverse pullback metric and on a regressor projected onto the tracking error and filtered torque-prediction error~\cite{Lee2018}, the resulting information gain is nonlinear and state-dependent. This illustrates why excitation cannot be characterized from the reference alone and motivates propagating covariance through the plant, controller, and adaptation law.
    
    The right-hand panels of Fig.~\ref{fig:fr3_3d_plot} show the resulting release positions. The conventional reference identifies the variation in CoM across the rods but not its absolute value: regressing the estimated CoM against the true value gives a slope of $0.92$ and an offset of $24$~mm, exceeding the $15$~mm tower half-width. The release positions consequently form a tight but displaced cluster, causing almost all rods to tip in the same direction. RO and OL reduce this offset to $4$ and $3$~mm, respectively, centering the release positions on the support. Fig.~\ref{fig:joint_plot} shows how the additional excitation is introduced. RO nearly doubles the total joint travel of the conventional reference ($10.9$ versus $6.1$~rad), while OL adds little total motion but redistributes it, with pronounced early motion in $q_5$. Both references concentrate additional motion in informative parts of the trajectory before returning to the common guarded descent for placement.
    
\begin{figure}[t]
        \centering
        \begin{subfigure}[t]{0.49\linewidth}
            \centering
            \includegraphics[width=\linewidth]{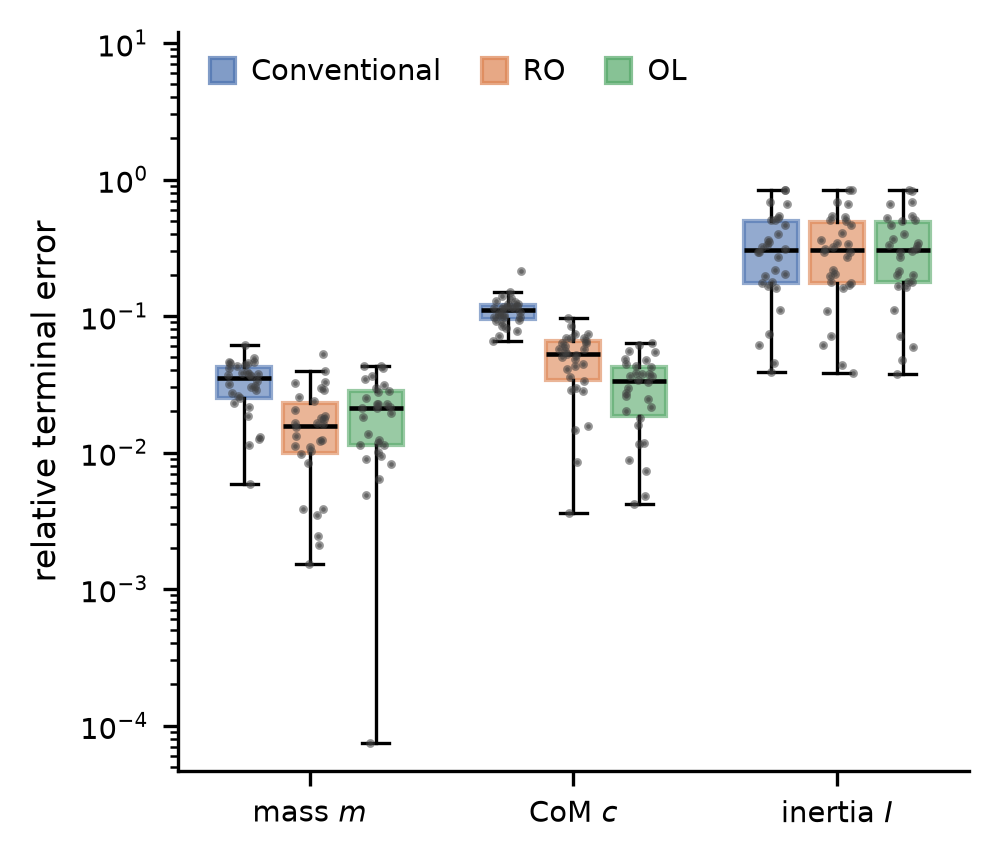}
            \caption{Terminal estimation error.}
            \label{fig:rod_paramclass}
        \end{subfigure}
        \hfill
        \begin{subfigure}[t]{0.49\linewidth}
            \centering
            \includegraphics[width=\linewidth]{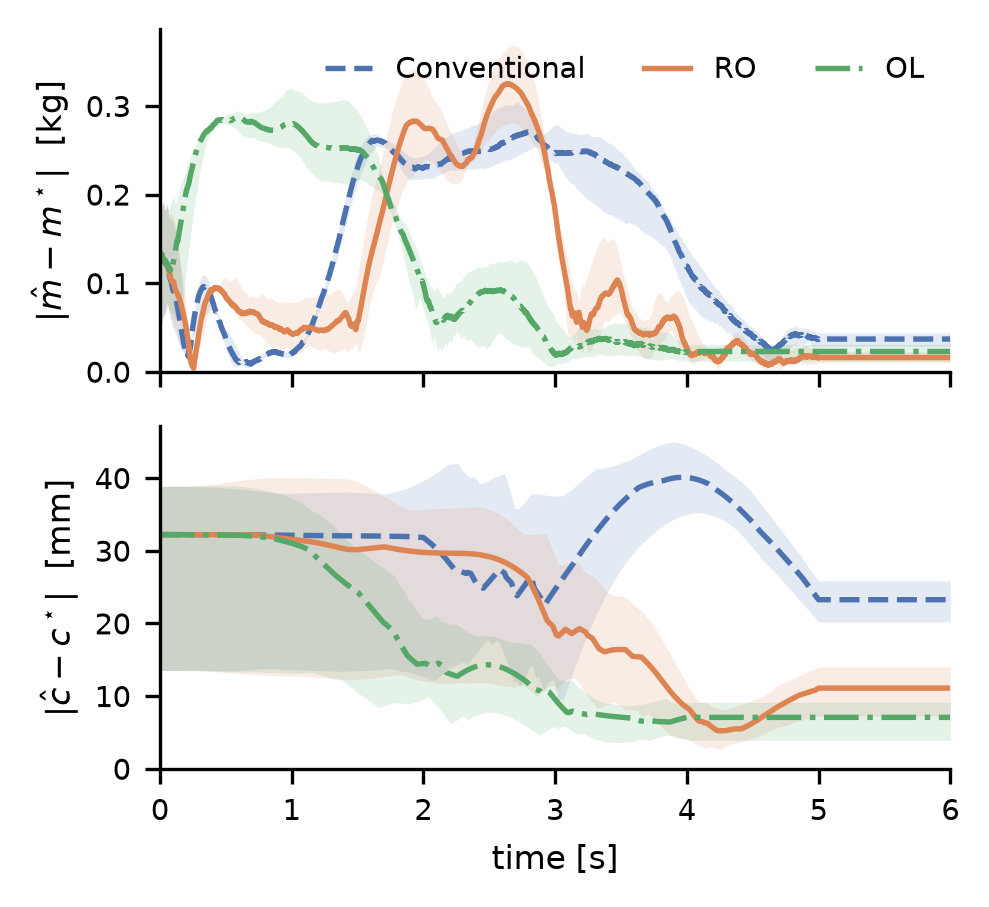}
            \caption{Estimate convergence.}
            \label{fig:rod_convergence}
        \end{subfigure}
        
        \caption{
        \textbf{(a)} Relative terminal error grouped by parameter class. \textbf{(b)} Median absolute error of the mass and CoM estimates over the $30$ rods.}
        \label{fig:rod_identification}
    \end{figure}

    Fig.~\ref{fig:rod_identification} further shows the selectivity of the resulting excitation. Both dual control references reduce CoM error throughout transport, whereas the conventional reference improves initially before degrading during the final approach. Mass identification is less monotonic because transport primarily excites the first moment $mc$, allowing mass and CoM to trade off until the motion separates them. At the terminal time, mass and CoM errors are approximately two and three times smaller for the dual control references, respectively, while the inertia tensor remains at approximately $30\%$ relative error for all methods. This is consistent with the near-singularity of $D$ in the inertia directions (Remark~2): resolving the inertia would require persistent, task-specific excitation that provides little benefit to the placement objective.

    \begin{table}[t]\centering
    \caption{Rod balancing in simulation, $30$ rods drawn from the prior. Mean $\pm$ standard deviation.}
    \label{tab:rod_sim}
    \begin{tabular}{lcccc}
    \toprule
     & balanced & $|\Delta m|/m$ & $|\Delta c|$ & placement \\
     & [-] & [--] & [mm] & error [mm] \\
    \midrule
    Conventional & $5/30$  & $0.033 \pm 0.013$ & $23.5 \pm 5.7$ & $15.2 \pm 0.9$ \\
    RO           & $27/30$  & $0.017 \pm 0.012$ & $10.7 \pm 4.9$ & $10.9 \pm 3.7$ \\
    OL           & $30/30$ & $0.021 \pm 0.012$ & $6.7 \pm 3.6$  & $8.1 \pm 3.1$ \\
    \bottomrule\end{tabular}
    \vspace{0.5em}
    \raggedright\footnotesize
    $|\Delta c|$ is the terminal CoM error, the parameter dimension the task depends on. Placement error is taken over the balanced runs only, since a rod that tips ends up on the table.
    \end{table}

\begin{figure}[t]
        \centering
        \begin{subfigure}[t]{0.48\linewidth}
            \centering \includegraphics[width=\linewidth]{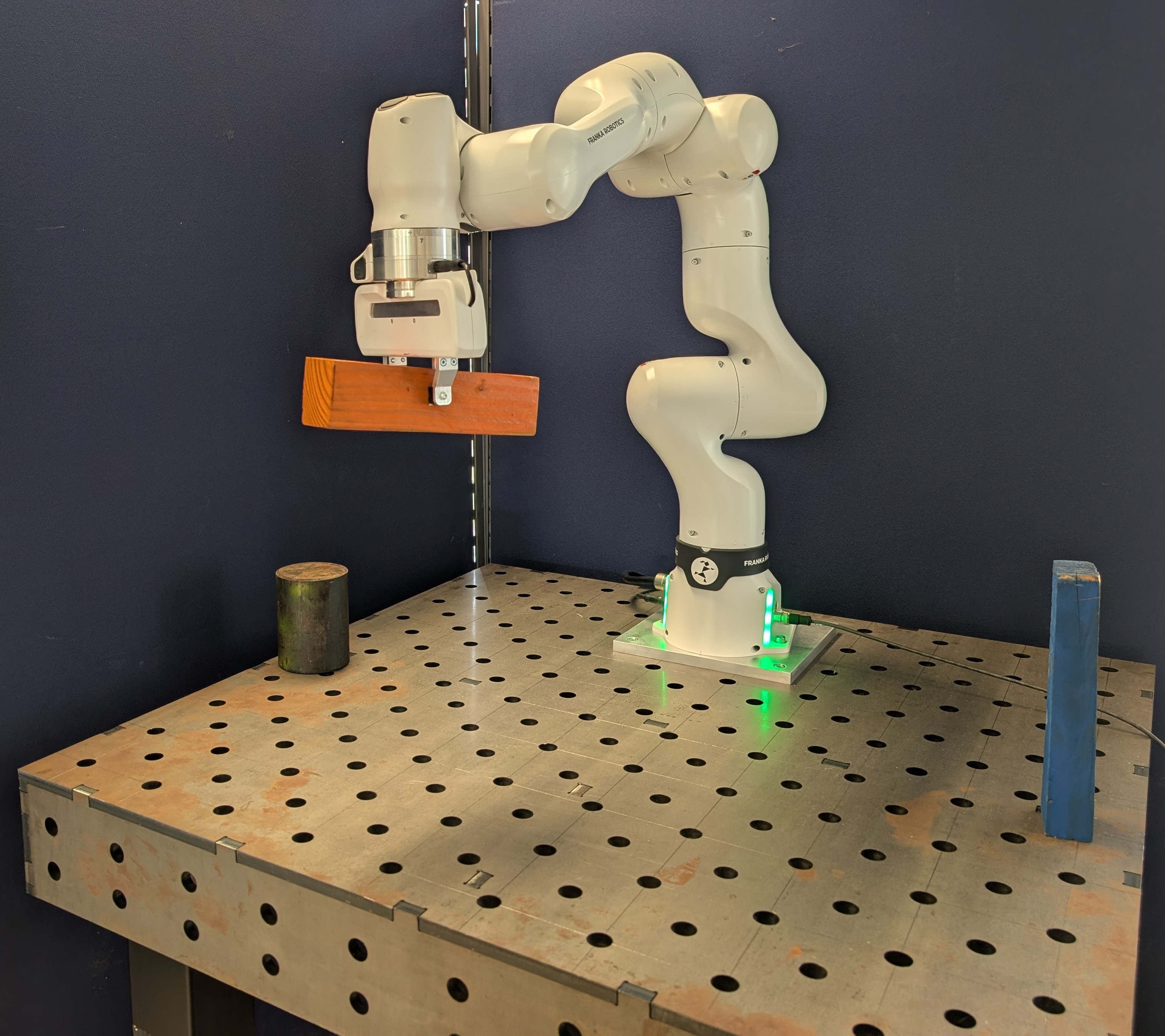}
        \end{subfigure}
        \hfill
        \begin{subfigure}[t]{0.48\linewidth}
            \centering \includegraphics[width=\linewidth]{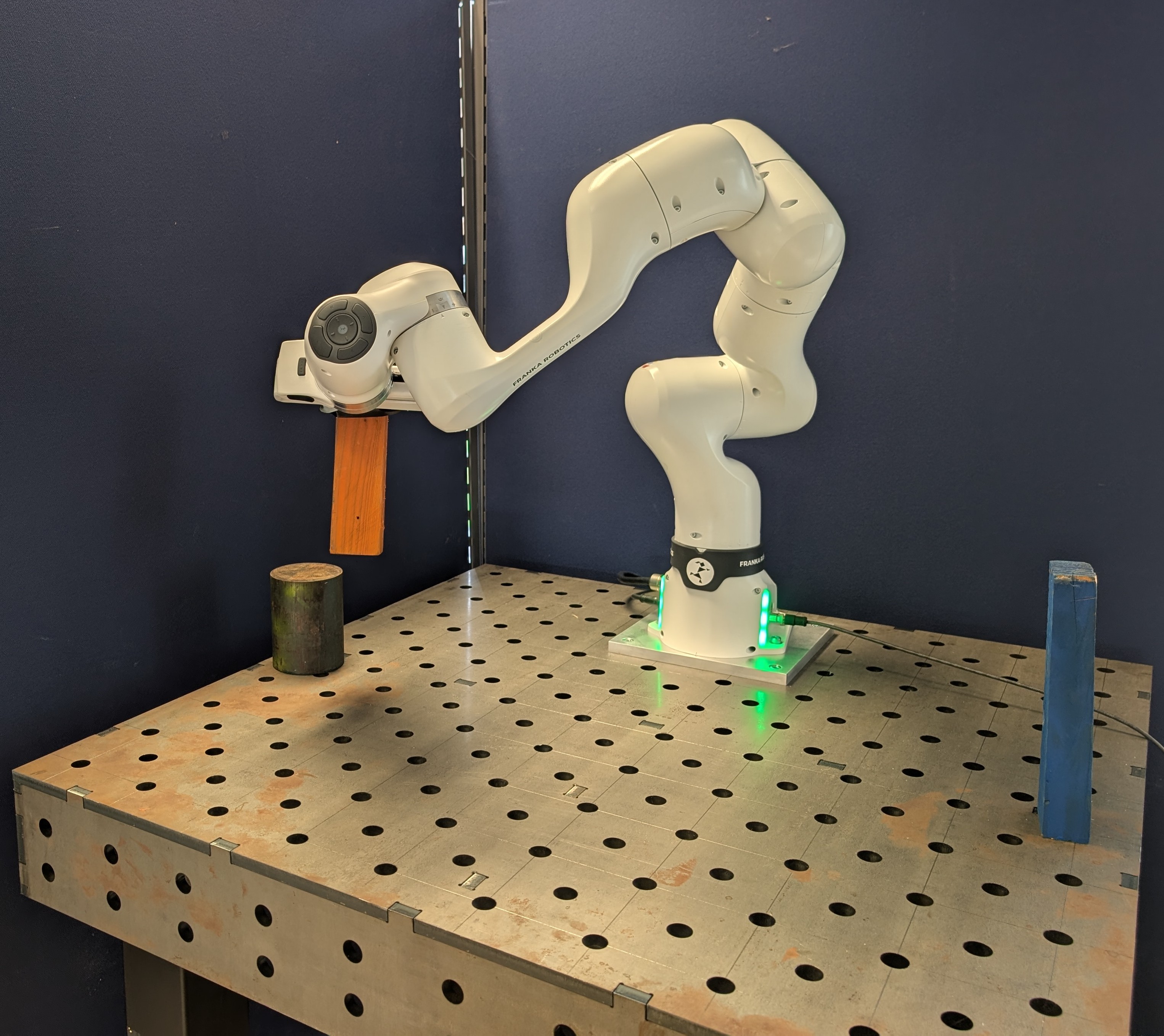}
        \end{subfigure}\\[1.2em]
        \begin{subfigure}[t]{0.48\linewidth}
            \centering \includegraphics[width=\linewidth]{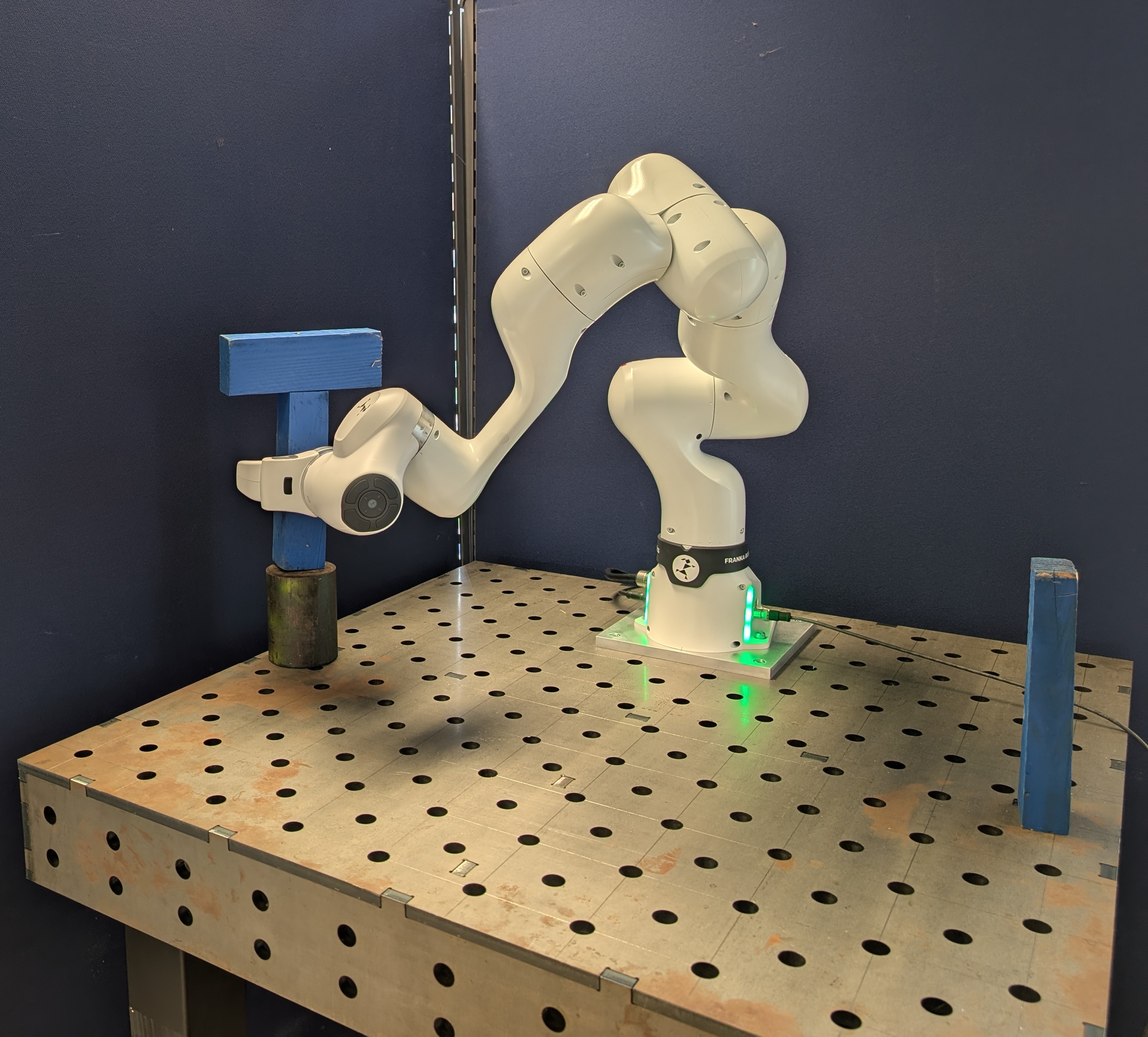}
        \end{subfigure}
        \hfill
        \begin{subfigure}[t]{0.48\linewidth}
            \centering \includegraphics[width=\linewidth]{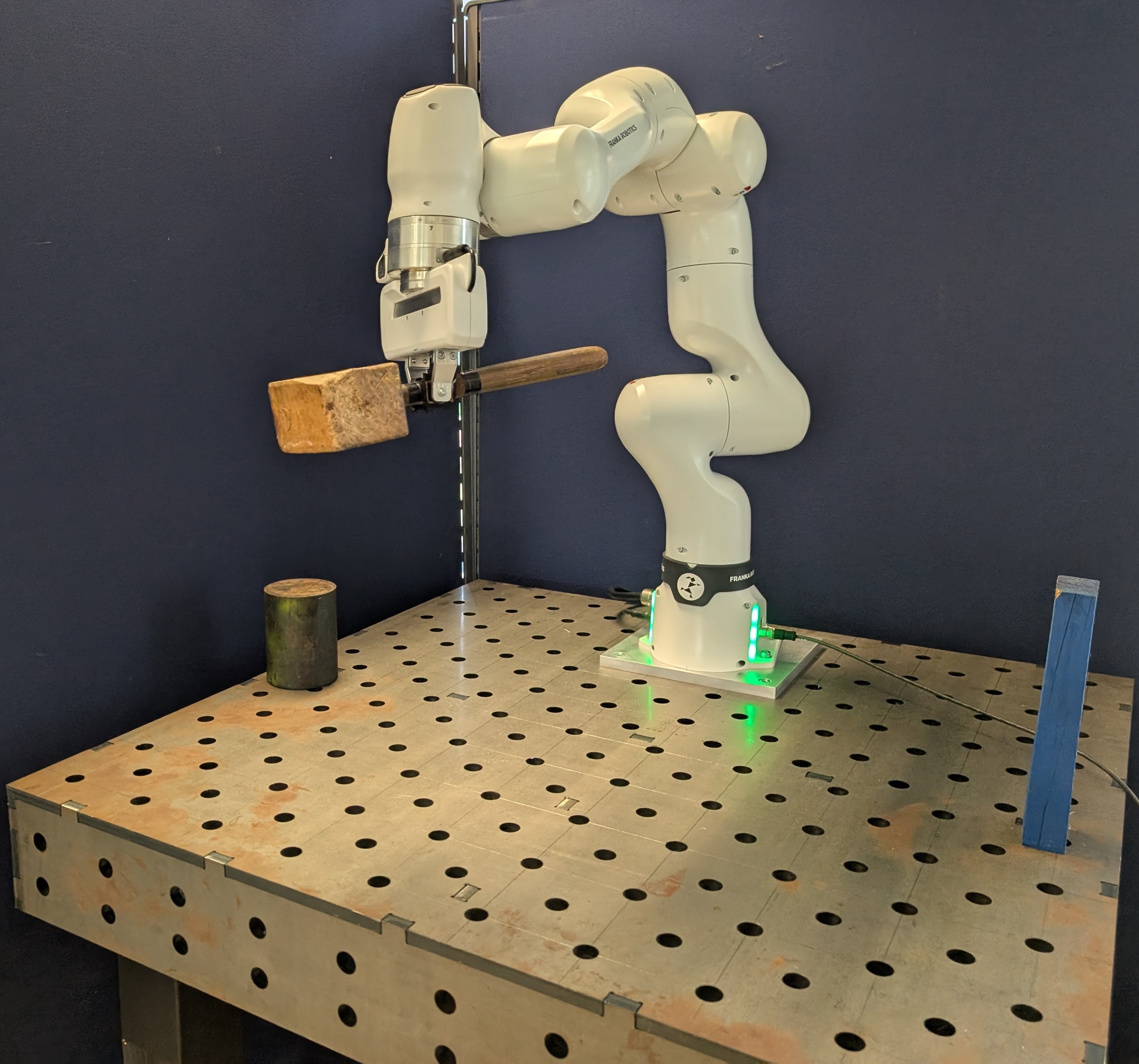}
        \end{subfigure}
        \caption{Hardware setup. The Franka FR3 grasps one of four test objects ---
        a wooden rod at its center, the same rod $7$\,cm off center, a T-shaped
        assembly, and a hammer --- transports it across the workspace and releases
        it onto the tower. None of the objects was used when designing the
        references.}
        \label{fig:fr3_hardware}
    \end{figure}

    \emph{Hardware.}
    We validate the approach on the physical FR3 by executing the simulation references unchanged: they are neither re-optimized nor re-tuned and were designed for a simulation prior corresponding to a $20$~cm beam with a $3\times3$~cm section and mean mass $1.1$~kg. The four test objects deliberately deviate from this prior (Fig.~\ref{fig:fr3_hardware}): a wooden rod ($447$~g, $24.5$~cm, $5\times5$~cm section) grasped at its center, the same rod grasped $7$~cm off center, a T-shaped assembly ($742$~g, $31$~cm), and a hammer ($775$~g, $34$~cm). All are lighter than the prior mean and longer than the nominal beam; the rod lies more than three prior standard deviations below the mean mass and below every rod sampled in simulation, while the T-shape and hammer lie outside the beam-and-lump family used to construct the prior. The hardware experiment therefore also tests robustness to a misspecified prior. The tower is identical to simulation. Ground-truth mass and approximate CoM are available, but the full inertia tensor is not, so we report binary success over $5$ repetitions per object (Table~\ref{tab:rod_hw}).
    
    The ordering observed in simulation is preserved. The conventional reference succeeds only for the centered rod ($4/5$), where the CoM coincides with the grasp and little identification is required, and fails in every repetition on the three asymmetric objects. RO and OL succeed on nearly every rod trial and most T-shape and hammer trials, achieving $15/20$ and $17/20$ successes, respectively, compared with $4/20$ for the conventional reference. The failures of the dual control references are concentrated on the two objects furthest from the training model class. That the same designed excitation transfers to these previously unseen objects suggests that it targets the task-relevant CoM direction rather than parameter values specific to the prior.

    \begin{table}[t]\centering
    \caption{Rod balancing on hardware: successful placements out of $5$ repetitions per object.}
    \label{tab:rod_hw}
    \begin{tabular}{lcccc|c}
    \toprule
     & rod & rod & T-shape & hammer & Overall \\
     & (centered) & ($7$\,cm off) & & \\
    \midrule
    Conventional & 4/5 & 0/5 & 0/5 & 0/5 & 4/20\\
    RO           & 5/5 & 4/5 & 3/5 & 3/5 & 15/20 \\
    OL           & 5/5 & 5/5 & 4/5 & 3/5 & 17/20 \\
    \bottomrule\end{tabular}\end{table}

    \section{Conclusion}\label{sec:conclusion}
    We presented two formulations for dual control reference generation: minimizing the expected task cost under parametric uncertainty, and minimizing the optimality loss incurred by planning with an incorrect estimate. Both propagate the prior through the closed loop formed by the plant, the feedback policy and its adaptation law, and both reduce to a nominal task cost augmented by a curvature-weighted parameter covariance. Fisher information thus arises from their construction rather than from an added exploration term, and is weighted by the task, so the design resolves only the parameter directions the task is sensitive to. Our results demonstrate that the proposed framework enhances the effectiveness of adaptive controllers by balancing exploration for parameter identification with exploitation for task execution.

    \appendices
    \section{Implementation Details}
    \label{app:impl}
    
    All three designs are implemented in JAX, with the closed-loop rollout differentiated end-to-end through the optimizer. The expectation in (\ref{eq:stoch}) is evaluated on a fixed deterministic set of samples from the prior, using stratified quantiles or $2L+1$ sigma points for a prior with $L$ latent variables. This makes the design objective deterministic during optimization. To improve convergence, the prior width is increased in stages, with each stage warm-started from the preceding solution. State and input constraints are enforced using an augmented Lagrangian. Platform-specific settings are summarized in Table~\ref{tab:impl}.
    
    \begin{table*}[t]
    \centering
    \footnotesize
    \caption{Implementation settings for the three experimental platforms.}
    \label{tab:impl}
    \begin{tabular}{@{}p{2.3cm}p{4.5cm}p{4.6cm}p{4.9cm}@{}}
    \toprule
     & Cartpole & Quadrotor with slung load & Franka FR3 \\
    \midrule
    Parameters $\theta$
    & $[m,h,J]$
    & $[m_L, L]$
    & $[m,c,I]\in\mathbb{R}^{10}$ \\
    
    Prior
    & $m\sim\mathcal{N}(0.30,0.40^2)$~kg; $h,J$ follow $m$
    & $m_L\sim\mathcal{N}(1.00,0.15^2)$~kg,
      $L\sim\mathcal{N}(0.70,0.20^2)$~m
    & beam $m_b\sim\mathcal{N}(0.65,0.10^2)$~kg,
      lump $m_p\sim\mathcal{N}(0.45,0.17^2)$~kg at
      $d\sim\mathcal{U}(\pm0.135)$~m \\
    
    Horizon
    & $10$~s
    & $24$~s
    & $5$~s + $1$~s guarded descent \\
    
    Reference
    & direct collocation, $501$ nodes
    & degree-5 B-spline on load path, $16$ control points
    & degree-5 joint-space B-spline, $12$ control points/joint \\
    
    Policy $\pi$
    & TVLQR, $Q=\mathrm{diag}(1,5,0.1,1)$, $R=1$
    & geometric SE(3)~\cite{lee2010geometric},
      $k_x=6$, $k_v=4$, $k_R=18$, $k_\Omega=3$
    & Slotine--Li, $\Lambda=8I$~s$^{-1}$,
      $K=\mathrm{diag}(\hat M)\Lambda$ \\
    
    Adaptation $\rho$
    & RLS, $\lambda=1$
    & RLS, $\lambda=1$
    & composite natural gradient~\cite{Lee2018},
      $\gamma=1$, $\kappa=0.3$ \\
    
    Rates
    & $50$~Hz
    & $250$~Hz
    & $1$~kHz \\
    
    Evaluation
    & MuJoCo, $15$ pole masses
    & MuJoCo, slung-load model, $30$ draws
    & MuJoCo, $30$ rods with shared noise seed $+$ hardware setup \\
    
    \bottomrule
    \end{tabular}
    \end{table*}

    \subsection{Measurement and Noise Models}
    \label{app:meas}
    
    All estimators use the parameter-linear residual
    \begin{equation*}
      r = Y(x,u)\theta+\varepsilon,
      \qquad
      \varepsilon\sim\mathcal{N}(0,R),
      \label{eq:app_measmodel}
    \end{equation*}
    with the regressor $Y$ constructed from measured states and inputs. Thus, measurement noise affects both the residual and the regressor. The resulting errors-in-variables model is approximated by incorporating the regressor noise into $R$. The platform-specific models are given below.
    
    \paragraph{Cartpole}
    The residual consists of the equations of motion with the known cart-mass contribution moved to the left,
    \[
    r=[\,F-m_c\ddot q_x,\;0\,]^{\!\top},
    \]
    with $Y$ constructed from the measured accelerations, velocities, and pole angle. The acceleration noise is $\sigma_{\ddot q}=10^{-2}$ and $\Sigma_0=\operatorname{diag}(0.05,0.02,0.005)$
    
    \paragraph{Quadrotor}
    The residual is the three-dimensional Newton equation of the payload in the world frame. It combines measured vehicle acceleration ($\sigma=0.30$~m/s$^2$) with cable angular acceleration obtained by finite differencing ($\sigma=0.25$~rad/s$^2$), using $R=I_3$. The $\beta$ column of $Y$ is proportional to the cable angular acceleration and therefore vanishes in hover. Errors in the regressor consequently have their strongest effect on the cable-length estimate, which is also the parameter most relevant to the delivery task.
    
    \paragraph{Franka FR3}
    The composite adaptation law does not explicitly use a measurement covariance. For the OL design, however, the measurement model determines the Fisher information. The innovation is $\tau_{\mathrm{meas}} -\tau(\hat\theta,q,\dot q,\ddot q)$  with sensor noise $\sigma_q=10^{-4}\ {\rm rad},\qquad \sigma_{\dot q}=2\times10^{-3}\ {\rm rad/s},\qquad \sigma_{\ddot q}=0.3\ {\rm rad/s^2}, \qquad \sigma_\tau=0.2\ {\rm N\,m}$ at the $1$~kHz control rate.
    
    The effective covariance used for the information calculation is \[ R_{\mathrm{eff}} = \frac{R_s}{N} +\eta R_m(\dot q,\ddot q)\] where $N=20$ is the number of samples in each estimator window. The white sensor-noise contribution $R_s/N$ therefore decreases with averaging, whereas the model-error term $R_m$, which increases with joint velocity and acceleration, is temporally correlated and is inflated by     $\eta=\tau_c/\Delta t\approx8.5$. The model-error covariance was identified on the physical robot. This weighting discounts information obtained during aggressive motion. The RO design does not use a measurement-noise model and    instead propagates covariance through the closed-loop sensitivities.

	\bibliographystyle{unsrt}
	\bibliography{references}

\end{document}